\documentclass[graybox]{svmult}

\usepackage{mathptmx}       
\usepackage{helvet}         
\usepackage{courier}        
\usepackage{type1cm}        
\usepackage{makeidx}         
\usepackage{graphicx}        
\usepackage{multicol}        
\usepackage[bottom]{footmisc}
\usepackage{subfigure}
\usepackage{mathtools}

\usepackage{amsmath}
\usepackage{amssymb}
\usepackage{url}
\usepackage{multirow}
\usepackage{tabularx}
\usepackage[disable]{todonotes}

\usepackage{enumitem} 
\usepackage{pbox} 

\usepackage[authoryear]{natbib}

\usepackage[Algorithm,chapter]{algorithm}
\usepackage[noend]{algpseudocode}

\usepackage{listings}
\lstdefinestyle{rm}{basicstyle={\fontsize{8.5pt}{8.5pt}\ttfamily},breaklines=true,keywordstyle=\slshape,language=[LaTeX]{TeX}, basewidth=0.6em,
    moredelim=**[is][\color{orange}]{@@}{@@}, xleftmargin=10pt, xrightmargin=10pt, belowskip=1.0em, aboveskip=1.0em,
}

\usepackage{tikz}
\usetikzlibrary{calc,arrows.meta,shapes}

\newcommand{\owncite}[1]{\citep{#1}}

\newcommand{\fullrule}{\par\noindent\rule{\textwidth}{0.7pt}\par\vspace{0.9pt}}

\usepackage{dcolumn}
\newcolumntype{.}{D{.}{.}{-1}}

\usepackage[pdftex]{hyperref}

\newboolean{comments}
\setboolean{comments}{false}
\newcommand{\comment}[2]{\ifthenelse{\boolean{comments}}{{\leavevmode\color{blue}\textbf{#1}:\textsl{#2}}}{}}
\newcommand{\TODO}[1]{}

\newcommand{\com}[1]{\todo[color=green!40,inline]{#1}}

\newcommand{\raus}[1]{}
\newcommand{\rauszwei}[1]{}

\newcommand{\labelfont}[1]{\emph{#1}}
\newcommand{\labelfontinv}[1]{\ensuremath{\overline{\text{\labelfont{#1}}}}}

\newcommand{\heur}{\ensuremath{h}}

\newcommand{\mvert}{\ensuremath{\,\big|\;}}  

\newcommand{\vektor}[1]{\ensuremath{\mathrm{\mathbf{#1}}}} 

\newcommand{\skalar}[1]{\ensuremath{\mathnormal{#1}}} 
\newcommand{\konstante}[1]{\ensuremath{\mathnormal{#1}}} 

\newcommand{\menge}[1]{\ensuremath{\mathcal{\uppercase{#1}}}}

\newcommand{\indexunten}[1][]{%
    \ensuremath{%
        \ifthenelse{\equal{#1}{\empty}}
            {}
            {_{#1}}%
    }%
}%
\newcommand{\indexoben}[1][]{%
    \ensuremath{%
        \ifthenelse{\equal{#1}{\empty}}
            {}
            {^{#1}}%
    }%
}%

\newcommand{\ds}[1]{\textsc{#1}} 

\newcommand{\docs}{\menge{X}}   
\newcommand{\doc}[1][]{\vektor{x}\indexunten[#1]} 
\newcommand{\doci}[1][]{\skalar{x}\indexunten[#1]} 
\newcommand{\natts}{\konstante{a}} 

\newcommand{\cats}{\menge{L}} 
\newcommand{\catsv}{\menge{Y}}
\newcommand{\catsi}[1][]{P\indexunten[#1]}  
\newcommand{\catsiv}[1][]{\vektor{y}\indexunten[#1]} 
\newcommand{\cat}[1][]{\skalar{\lambda}\indexunten[#1]}
\newcommand{\cati}[1][]{\skalar{\lambda}\indexunten[#1]}
\newcommand{\catv}[1][]{\skalar{y}\indexunten[#1]}
\newcommand{\ncats}{\konstante{n}} 
\newcommand{\ndocs}{\konstante{m}} 
\newcommand{\docset}{\menge{T}}
\newcommand{\trainset}{\docset}

\newcommand{\classifierBin}{f}  
\newcommand{\classifier}{\classifierBin}

\newlength{\lueckehelp}%
\newcommand{\luecke}[1]{%
\settowidth{\lueckehelp}{#1}%
\hspace{\lueckehelp}%
}%
\newcommand{\learner}[1]{\textsl{#1}}

\newcommand{\fea}{\phi} 

\newcommand{\rul}{\ruln{r}} 
\newcommand{\ruln}[1]{\mathbf{#1}}
\newcommand{\sRul}{R} 

\newcommand{\svalue}[1]{\textsf{\mbox{#1}}}
\newcommand{\true}{\svalue{true}}

\long\def\symbolfootnotetext[#1]#2{\begingroup%
\def\thefootnote{\fnsymbol{footnote}}\footnotetext[#1]{#2}\endgroup}

\newcommand{\ripper}{\learner{Ripper}}

\rm

\makeindex             

\begin{document}

\title*{Learning Interpretable Rules  \\
	for Multi-label Classification \\
	\normalsize \normalfont \ \\Preprint version. To be published in: Explainable and Interpretable Models in Computer Vision and Machine Learning. The Springer Series on Challenges in Machine Learning. Springer (2018)}
\titlerunning{Learning Interpretable Rules for Multi-label Classification}

\author{Eneldo Loza Menc\'ia , Johannes F\"urnkranz,  Eyke H\"ullermeier,  Michael Rapp}
\institute{ Eneldo Loza Menc\'ia \at Technische Universit\"at Darmstadt, Knowledge Engineering Group,
\\ \mbox{\email{eneldo@ke.tu-darmstadt.de}}
\and
Johannes F\"urnkranz \at Technische Universit\"at Darmstadt, Knowledge Engineering Group,
\\ \mbox{\email{juffi@ke.tu-darmstadt.de}}
 \and
Eyke H\"ullermeier \at Universit\"at Paderborn, Intelligent Systems,
\email{eyke@upb.de}
\and
Michael Rapp\at Technische Universit\"at Darmstadt, Knowledge Engineering Group,
\\ \mbox{\email{mrapp@ke.tu-darmstadt.de}}
}

\maketitle

\abstract{
Multi-label classification (MLC) is a supervised learning problem in which, contrary to standard multiclass classification, an instance can be associated with several class labels simultaneously. In this chapter, we advocate a rule-based approach to multi-label classification. Rule learning algorithms are often employed when one is not only interested in accurate predictions, but also requires an interpretable theory that can be understood, analyzed, and qualitatively evaluated by domain experts. Ideally, by revealing patterns and regularities contained in the data, a rule-based theory yields new insights in the application domain. Recently, several authors have started to investigate how rule-based models can be used for modeling multi-label data. Discussing this task in detail, we highlight some of the problems that make rule learning considerably more challenging for MLC than for  conventional classification. While mainly focusing on our own previous work, we also provide a short overview of related work in this area.
}

\section{Introduction}



Multi-label classification (MLC) is a problem in the realm of supervised learning. Contrary to conventional, single-label classification, MLC allows an instance to be associated with multiple class labels simultaneously. Dealing with and taking advantage of (statistical) dependencies between the presence and absence (relevance and irrelevance) of different labels has been identified as a key issue in previous work on MLC. To improve predictive performance, essentially all state-of-the-art MLC algorithms therefore seek to capture label dependencies in one way or the other.

In this chapter, we will argue that inductive rule learning is a promising approach for tackling MLC problems. In particular, rules provide an interpretable model for mapping inputs to outputs, and allow for tightly integrating input variables and labels into coherent comprehensible theories.
%
%
For example, so-called global dependencies between labels can be explicitly modeled and expressed in the form of rules. Moreover, 
these can be easily generalized to local dependencies, which include regular input features as a local context in which a label dependency holds.
Such rules, which mix labels and features, are nevertheless directly interpretable and comprehensible for humans. Even if complex and long rules are generated, the implication between labels can be easily grasped 
by focusing on the part of the rules that actually considers the labels.
Hence, in contrast to many other model types that capture label dependencies implicitly, such dependencies can be analyzed and interpreted more directly.

We will start with a brief definition and formalization of the multi-label learning problem (Section~\ref{sec:MLC}), in which we also introduce a dataset that will serve as a running example. In Section~\ref{section_multi_label_rules}, we then define multi-label rules, highlighting the differences to conventional classification rules, discuss various dimensions and choices that have to be made, and list some challenges for learning such rules. Sections~\ref{sec:discovery} and~\ref{sub:stacking} then deal with descriptive and predictive multi-label rule learning, respectively. The former recalls association-rule based approaches and discusses how properties like anti-monotonicity can be used to efficiently search for a suitable head for a given rule body, whereas the latter discusses two approaches for learning predictive rule-based theories: one based on stacking different label prediction layers, and another one based on adapting the separate-and-conquer or covering strategy to the multi-label case. Finally, in Section~\ref{sec:eval}, we present and discuss rule-based theories for a few well-known sample multi-label databases, before we conclude in Section~\ref{sec:conclusions}.



\section{Multi-Label Classification}
\label{sec:MLC}

Multi-label classification has received a lot of attention in the recent machine learning
literature
\citep{tsoumakas10MLoverview,Multilabel-SI-MLJ,ML:Review,Multilabel-Tutorial,XC-15,Multilabel-Book,ZhangZ14}.
The motivation for MLC originated in the field of text categorization \citep{hayes90construe,Lewis:SIGIR-92,Reuters-21578}, but nowadays multi-label methods are used in
applications as diverse as
music categorization \citep{trohidis08emotions},
semantic scene classification \citep{Scene-Data},
or protein function classification \citep{RankSVM}.

\subsection{Problem Definition}

The task of MLC is to associate an instance with one or several labels $\lambda_{i}$ out of a finite label space $\cats$.
with $\ncats = |\cats|$ being the number of available labels.
Contrary to ordinary classification, MLC allows each instance to be associated with more than one (class) label, but,
in contrast to multiclass learning, alternatives
are not assumed to be mutually exclusive, such that multiple labels
may be associated with a single instance.
Figure~\ref{tab:ML-set} shows an example, which relates persons described with some demographic characteristics to the newspapers and magazines they subscribe. Obviously, the number of subscriptions can vary.
For example, subject \#1 (a single male with primary education and no kids) has subscribed to no magazines at all, whereas \#13 (a divorced male with university degree and children) obtains a quality newspaper and a tabloid.

\begin{figure}[t!]
	\begin{center}
\subfigure[With set-valued outputs\label{tab:ML-set}]{%
\begin{tabular}{|c|cccc|c|}
  	\hline
    \multicolumn{5}{|c|}{\bf Person} & \textbf{Subscribed Magazines}\\
\textbf{No.} &
\textbf{Education}&
\textbf{Marital} &
\textbf{Sex} &
\textbf{Children?} &
\\
\hline \hline
1 &
Primary &
Single &
Male &
No &
$\emptyset$ \\
2 &
Primary &
Single &
Male &
Yes &
$\emptyset$\\
3 &
Primary &
Married &
Male &
No &
\{tabloid\}\\
4 &
University &
Divorced &
Female &
No &
\{quality, fashion\}\\
5 &
University &
Married &
Female &
Yes &
\{quality, fashion\}\\
6 &
Secondary &
Single &
Male &
No &
\{tabloid\}\\
7 &
University &
Single &
Male &
No &
\{quality, tabloid\}\\
8 &
Secondary &
Divorced &
Female &
No &
\{quality, sports\}\\
9 &
Secondary &
Single  &
Female &
Yes &
\{tabloid, fashion\}\\
10 &
Secondary &
Married &
Male &
Yes &
\{quality, tabloid\}\\
11 &
Primary &
Married &
Female &
No &
$\emptyset$\\
12 &
Secondary &
Divorced &
Male &
Yes &
$\emptyset$\\
13 &
University &
Divorced &
Male &
Yes &
\{quality, tabloid\}\\
14 &
Secondary &
Divorced &
Male &
No &
\{quality, sports\}\\
\hline
        \end{tabular}
}

\subfigure[With binary output vectors\label{tab:ML-binary}]{
		\begin{tabular}{|c|cccc|cccc|}
			\hline
			\multicolumn{5}{|c|}{\bf Person} & \multicolumn{4}{c|}{\textbf{Subscribed Magazines}}\\
			\textbf{No.} &
			\textbf{Education}&
			\textbf{Marital} &
			\textbf{Sex} &
			\textbf{Children?} &
			\textbf{Quality} &
			\textbf{Tabloid} &
			\textbf{Fashion} &
			\textbf{Sports} \\
			\hline \hline
			1 &
			Primary &
			Single &
			Male &
			No &
			0 &
			0 &
			0 &
			0 \\
			2 &
			Primary &
			Single &
			Male &
			Yes &
			0 &
			0 &
			0 &
			0 \\
			3 &
			Primary &
			Married &
			Male &
			No &
			0 &
			1 &
			0 &
			0 \\
			4 &
			University &
			Divorced &
			Female &
			No &
			1 &
			0 &
			1 &
			0 \\
			5 &
			University &
			Married &
			Female &
			Yes &
			1 &
			0 &
			1 &
			0 \\
			6 &
			Secondary &
			Single &
			Male &
			No &
			0 &
			1 &
			0 &
			0 \\
			7 &
			University &
			Single &
			Male &
			No &
			1 &
			1 &
			0 &
			0 \\
			8 &
			Secondary &
			Divorced &
			Female &
			No &
			1 &
			0 &
			0 &
			1 \\
			9 &
			Secondary &
			Single  &
			Female &
			Yes &
			0 &
			1 &
			1 &
			0 \\
			10 &
			Secondary &
			Married &
			Male &
			Yes &
			1 &
			1 &
			0 &
			0 \\
			11 &
			Primary &
			Married &
			Female &
			No &
			0 &
			0 &
			0 &
			0 \\
			12 &
			Secondary &
			Divorced &
			Male &
			Yes &
			0 &
			0 &
			0 &
			0 \\
			13 &
			University &
			Divorced &
			Male &
			Yes &
			1 &
			1 &
		0 &
			0 \\
			14 &
			Secondary &
			Divorced &
			Male &
			No &
			1 &
			0 &
			0 &
			1 \\
			\hline
		\end{tabular}
}
\end{center}
	\caption{Two representations of a sample multi-label classification problem, which relates demographic characteristics to subscribed newspapers and magazines.}\label{tab:ML-example}
\end{figure}

Potentially, there are $2^\ncats$ different allowed
allocations of
the output space,
which is a dramatic growth compared to the
$\ncats$ possible states in the multiclass setting.
%
However, not all possible combinations need to occur in the database. For example, nobody in this database has subscribed to both a fashion and a sports magazine.
Note that these label attributes are not independent. The fact
that there may be
correlations and dependencies between the labels in $\cats$ makes the multi-label setting particularly challenging and interesting compared to the classical setting of binary and multiclass classification.

Formally, MLC refers to the task of learning a predictor $\classifier:\, \docs \rightarrow 2^\cats$
that maps elements $\doc$
of an instance space $\docs$ to subsets $\catsi$ of a set of labels $\cats=\{\cat[1],\dots,\cat[\ncats]\}$. Equivalently, predictions can be expressed as binary vectors $\catsiv= \classifier(\doc) = (\catv[1],\ldots,\catv[\ncats]) \in \{0,1\}^\ncats$, where
each attribute \catv[i] encodes the  presence
($1$) or absence ($0$) of the corresponding label $\cat[i]$,
We will use these two notations interchangeably, i.e., $\catv$ will be used to refer to an element in a binary prediction vector, whereas $\cat$ refers to an element in a predicted label set.

An instance $\doc_{j}$ is in turn represented in attribute-value form, i.e., it consists of a vector
$\doc_{j} \coloneqq \langle \doci_{1},...,\doci_{\natts} \rangle \in \docs = \fea_{1} \times ... \times \fea_{\natts}$, where $\fea_i$ is a numeric or nominal attribute.

Consequently, the training data set of an MLC problem can be defined as a sequence of tuples $\trainset \coloneqq \langle (\doc_{1},\catsiv_{1}),...,(\doc_{\ndocs},\catsiv_{\ndocs}) \rangle \subseteq \docs \times \catsv$ with $\ndocs = |\trainset|$.
Figure~\ref{tab:ML-example} shows both representations, once with sets as outputs (\ref{tab:ML-set}) and once with binary vectors as outputs (\ref{tab:ML-binary}).

\subsection{Dependencies in Multi-Label Classification}

The simplest and best known approach to multi-label classification is
\emph{binary relevance} (BR) learning \citep[e.g.][]{tsoumakas10MLoverview}.
It tackles a multi-label problem by learning one classifier for each label, using all its instances as positive and the others as negative examples.
The obvious disadvantage of this transformation
is the ignorance of possible dependencies between the labels.
More advanced methods seek to exploit such dependencies,
mainly with the goal of improving predictive accuracy.




The goal of most classification algorithms is to capture dependencies between input variables $\doci[j]$ and the output variables $\catv_i$. In fact, the prediction $\hat{\catv}=\classifier(\doc)$ of a scoring classifier $\classifier$ is often regarded as an approximation of  the conditional probability $\Pr(\catv = \hat{\catv} \mvert \doc)$, i.e., the probability that $\hat{\catv}$ is the true label for the given instance $\doc$. In MLC, dependencies may not only exist between $\doc$ and each target $\catv_i$, but also between the labels $\catv_1, \ldots , \catv_\ncats$ themselves.

A key distinction is between
\emph{unconditional} and \emph{conditional independence} of labels.
In the first case, the joint distribution $\Pr(\catsiv)$ in the label space factorizes into the product of the marginals $\Pr(\catv_i)$, i.e., $\Pr(\catsiv) = \Pr(\catv_1)  \cdots \Pr(\catv_\ncats)$, whereas in the latter case, the factorization $\Pr(\catsiv \mvert \doc) = \Pr(\catv_1 \mvert \doc)  \cdots \Pr(\catv_\ncats \mvert \doc)$ holds conditioned on $\doc$, for every instance $\doc$.
In other words, unconditional dependence is a kind of global dependence (for example originating from a hierarchical structure on the labels), whereas conditional dependence is a dependence locally restricted to a single point in the instance space.

In the literature, both types of dependence have been explored. For example,
\citet{MLC-BayesNet} model label dependence in the form of a Bayesian network.
\citet{ML:ExploitingLabelDependencies} provide an empirical analysis of the different types of dependencies between pairs of labels on standard benchmark datasets, and analyze the usefulness of modeling them.
Unconditional dependencies were analyzed by a simple $\chi^2$ test on the label co-occurrence matrix, whereas for detecting unconditional dependencies they compared the performance of a classifier $\classifier_i$ for a label $\catv_i$ trained on the instance features $(\doc)$ to the same learning algorithm being applied to the input space $(\doc,\catv_j)$ augmented by the label feature of a second label $\catv_j$.
If the predictions differ statistically significantly, then $\catv_i$ is assumed to be conditionally dependent on $\catv_j$.
Their evaluations show that pairwise unconditional dependencies occur more frequently than pairwise conditional dependencies, and that, surprisingly, modeling global dependencies is more beneficial in terms of predictive performance. However, this finding is very specific to their setting, where the dependence information is basically used to guide a decomposition into smaller problems with less labels that are either independent or dependent. In addition, only pairwise co-occurrence and pairwise exclusion can effectively be exploited by their approach. As we will see in Section~\ref{sec:MLC-rules}, rules can be used to flexibly formulate a variety of different dependencies, including partially label-dependent or local dependencies.


\subsection{Evaluation of Multi-label Predictions}
\label{sec:evaluation}


\subsubsection{Bipartition Evaluation Functions}

To evaluate the quality of multi-label predictions, we use bipartition evaluation measures (cf.\ \citet{tsoumakas10MLoverview}) which are based on evaluating differences between true (\emph{ground truth}) and predicted label vectors. They  can be considered as functions of two-dimensional \emph{label confusion matrices} which represent the \emph{true positive} ($TP$), \emph{false positive} ($FP$), \emph{true negative} ($TN$) and \emph{false negative} ($FN$) label predictions. For a given example $\doc_{j}$ and a label $\catv_{i}$ the elements of an atomic confusion matrix $C_{i}^{j}$ are computed as
\begin{equation}
\label{equation_confusion_matrix}
C_i^j = \left(\begin{matrix}
TP_i^j & FP_i^j \\
FN_i^j & TN_i^j
\end{matrix}\right)
= \left( \begin{matrix}
\catv_{i}^j  \hat{\catv}_{i}^j \ && \
(1-\catv_{i}^j)  \hat{\catv}_{i}^j \\
(1-\catv_{i}^j)  (1-\hat{\catv}_{i}^j) \ && \
\catv_{i}^j  (1-\hat{\catv}_{i}^j)
\end{matrix}\right)
\end{equation}
where the variables $\catv_{i}^j$ and $\hat{\catv}_{i}^j$ denote the absence (0) or presence (1) of label $\lambda_{i}$ of example $\doc_j$ according to the ground truth or the predicted label vector, respectively.

Note that for candidate rule selection we assess $TP$, $FP$, $TN$, and $FN$ differently. To ensure that absent and present labels have the same impact on the performance of a rule, we always count correctly predicted labels as $TP$ and incorrect predictions as $FP$, respectively. Labels for which no prediction is made are counted as $TN$ if they are absent, or as $FN$ if they are present.

\subsubsection{Multi-Label Evaluation Functions}

In the following some of the most common bipartition metrics $\delta(C)$ used for MLC are presented (cf., e.g., \citet{tsoumakas10MLoverview}). They are mappings $\mathbb{N}^{2x2} \rightarrow \mathbb{R}$ that assign a real-valued score (often normalized to $[0,1]$) to a confusion matrix $C$. Predictions that reach a greater score outperform those with smaller values.
\begin{itemize}
	\item \textbf{Precision:} Percentage of correct predictions among all predicted labels.
	\begin{equation}
	\label{equation_precision}\footnotesize
	\delta_{prec}(C) \coloneqq \frac{TP}{TP+FP}
	\end{equation}
	\item \textbf{Hamming accuracy:} Percentage of correctly predicted present and absent labels among all labels.
	\begin{equation}
	\label{equation_hamming_accuracy}\footnotesize
	\delta_{hamm}(C) \coloneqq \frac{TP+TN}{TP+FP+TN+FN}
	\end{equation}
	\item \textbf{F-measure:} Weighted harmonic mean of precision and recall. If $\beta < 1$, precision has a greater impact. If $\beta > 1$, the F-measure becomes more recall-oriented.
	\begin{equation}
	\label{equation_f_measure}\footnotesize
	\delta_{F}(C) \coloneqq \frac{\beta^2 + 1}{\frac{\beta^2}{\delta_{rec}(C)} + \frac{1}{\delta_{prec}(C)}} \text{ , with } \delta_{rec}(C) = \frac{TP}{TP + FN} \text{ and } \beta \in \left[0,\infty\right]
	\end{equation}
	\item \textbf{Subset accuracy:} Percentage of perfectly predicted label vectors among all examples. Per definition, it is always calculated using example-based averaging.
	\begin{equation}
	\label{equation_subset_accuracy}\footnotesize
	\delta_{acc}(C) \coloneqq \frac{1}{m} \sum \limits_{j} \left[ \catsiv_{j} = \hat{\catsiv}_{j} \right] \text{ , with } [x] = \begin{cases}
	1, & \text{if $x$ is true} \\
	0, & \text{otherwise}
	\end{cases}
	\end{equation}
\end{itemize}

\subsubsection{Aggregation and Averaging}

When evaluating multi-label predictions that have been made for $m$ examples with $n$ labels, one has to deal with the question of how to aggregate the resulting $m \cdot n$ atomic confusion matrices. Essentially, there are four possible averaging strategies -- either \emph{(label- and example-based) micro-averaging}, \emph{label-based (macro-)averaging}, \emph{example-based (macro-) averaging} or \emph{(label- and example-based) macro-averaging}. Due to the space limitations, we restrict our analysis to the most popular aggregation strategy employed in the literature, namely \emph{micro-averaging}. This particular averaging strategy is formally defined as
\begin{equation}
\label{equation_micro_averaging}
\footnotesize
\delta(C) = \delta \left( \sum \nolimits_{j}^{} \sum \nolimits_{i}^{} C_{i}^j \right) \equiv \delta \left( \sum \nolimits_{i}^{} \sum \nolimits_{j}^{} C_{i}^j \right) \, ,
\end{equation}
where the $\sum$ operator denotes the cell-wise addition of confusion matrices.

\section{Multi-Label Rule Learning}
\label{section_multi_label_rules}

In this section, we discuss rule-based approaches for multi-label classification. We start with a brief recapitulation of inductive rule learning.

\subsection{Rule Learning}
\label{sec:rule-learning}

Rule learning has a very long history and is a well-known problem in the machine learning community \citep{jf:Book-Nada}. Over the years many different algorithms to learn a set of rules were introduced.
The main advantage of rule-based classifiers is the interpretability of the models
as rules can be more easily comprehended by humans than other models such as neural networks.
Also, it is easy to define a syntactic generality relation, which helps to structure the search space.
The structure of a rule offers the calculation of overlapping of rules as well as \emph{more specific} and \emph{more general}-relations. Thus, the rule set can be easily modified as opposed to most statistical models such as SVMs or neural networks. However, most rule learning algorithms are currently limited to binary or multiclass classification.
Depending on the goal, one may discriminate between predictive and descriptive approaches.


\subsubsection{Predictive Rule Learning}
\label{sec:predictive-rules}

\emph{Classification rules} are commonly expressed  in the form
\begin{equation}
\rul: H \leftarrow B,
\end{equation}
where the body $B$ consists of a number of \emph{conditions},
which are typically formed from
the attributes of the instance space,
and
the head $H$ simply assigns a value to the output attribute (e.g.,
$\catv=0 $ or $\catv=1$ in binary classification).
We refer to this type of rules as \emph{single-label head} rules.

For combining such rules into predictive theories, most algorithms
follow the \emph{covering} or \emph{separate-and-conquer}
strategy \citep{jf:AI-Review}, i.e., they proceed by learning one rule
at a time. After adding a rule to the growing rule set, all examples
covered by this rule are removed, and the next rule is learned from
the remaining examples.
In order to prevent overfitting, the two constraints that all examples have to be covered (\emph{completeness}) and that negative examples must not be covered (\emph{consistency}) can be relaxed so that some positive examples may remain uncovered and/or some negative examples may be covered by the set of rules.
Typically, heuristics are used that trade off these two objectives and guide the search towards solutions that excel according to both criteria \citep{jf:MLJ-Heuristics,AntMiner-Evaluation}.\footnote{Some algorithms, such as ENDER \citep{ENDER}, also
	find rules that directly minimize a regularized loss function.}
This may also be viewed as a simple instance of the \emph{LeGo}
framework for combining local patterns (individual rules)
to global theories (rule sets or decision lists)
\citep{jf:LeGo-08-WS-Paper}.
%

\subsubsection{Descriptive Rule Learning}
\label{sec:descriptive-rules}

Contrary to predictive approaches, descriptive rule learning algorithms typically focus on individual rules.
For example, \emph{subgroup discovery} algorithms
\citep{SupervisedDescriptiveRuleDiscovery} aim at discovering
groups of data that have an unusual class distribution, or \emph{exceptional
model mining} \citep{ExceptionalModelMining} generalizes this
notion to differences with respect to data models instead of data
distributions.
\citet{duivesteijn12MultilabelLego} extended the latter approach to MLC
for
finding local exceptionalities in the dependence relations between
labels.


Arguably the best-known descriptive approach are \emph{association rules}
\citep{FrequentPatternMining-Survey,AssociationRules-Overview,AssociationRules-Book}, which relate properties of the data in the body of the rule
to other properties in the head of the rule. Thus, contrary to
classification rules, where the head consists of a single class label, multiple conditions may appear in the head.
Typically, association rules are found by exhaustive search, i.e., all rules
that satisfy a minimum support and minimum confidence threshold are
found \citep{APriori,Eclat,FP-Growth}, and subsequently filtered and/or
ordered according to heuristics. Only few algorithms directly find
rules that optimize a given score function
\citep{AssociationRules-OPUS}.
They can also be combined into theories with class association rule
learning algorithms such as CBA \citep{CBA,CBA2,jf:LeGo-08-WS-Sulzmann}.

\subsection{Multi-Label Rules}
\label{sec:MLC-rules}


The goal of multi-label rule learning is to discover rules of the form
\begin{equation}\rul: \hat{\catsiv} \leftarrow B
\end{equation}
The head of the rule may be viewed as a binary prediction vector $\hat{\catsiv}$, or as a set of predicted labels $\hat{\catsi} \subset \cats$.
The body may consist of several conditions, which the examples that are covered by the rule have to satisfy. In this work, only conjunctive, propositional rules are considered, i.e., each condition compares an attribute's value to a constant by either using equality (nominal attributes) or inequalities (numerical attributes).

In mixed representation rules,
labels may occur both as rule features (in the body
of the rule) and as predictions (in the head of the rule).
Formally, we intend to learn rules of the form
\begin{equation}
\rul: \catv^{(i+j+k)}, \dots  , \catv^{(i+j+1)}
\leftarrow
\catv^{(i+j)},  \dots , \catv^{(i+i)},
\fea^{(i)}, \dots , \fea^{(1)}
\label{eq:rule}
\end{equation}
in which $i \geq 0$ Boolean features
, which characterize the input
instances, can be mixed with $j \geq 0$ labels
in the body of the rule, and are mapped to $k > 0$ different labels
in the head of the rule.

\begin{table}[t!]
	\caption{Examples of different forms of multi-label rules based on the sample dataset  Figure~\ref{tab:ML-example}. Attribute names in italic denote label attributes, attributes with an overline denote negated conditions.}
	\label{tab:rule_forms}
	\centering
	\resizebox{\textwidth}{!}{
		\begin{tabular}{|ll|l|l|}
			\hline
			\multicolumn{2}{|c|}{\textbf{Head}} & \multicolumn{1}{c|}{\textbf{Body}} & \multicolumn{1}{c|}{\textbf{Example rule}}\\
\hline\hline
\rule{0pt}{1.1em} 
			\multirow{2}{*}{Single-label} & Positive  & \multirow{2}{*}{Label-independent} &
                        \labelfont{quality}     $\leftarrow$     University, Female  \\
			& Negative & & \labelfontinv{tabloid} $\leftarrow$  Secondary, Divorced\\
			\hline
\rule{0pt}{1.1em} 
			\multirow{2}{*}{Single-label} & Positive & \multirow{2}{*}{Partially Label-dependent} & \labelfont{quality} $\leftarrow$ \labelfontinv{tabloid}, University\\
			& Negative & & \labelfontinv{quality} $\leftarrow$ \labelfontinv{tabloid}, Primary\\
			\hline \rule{0pt}{1.1em} 
			\multirow{2}{*}{Single-label} & Positive & \multirow{2}{*}{Fully Label-dependent} & \labelfontinv{sports} $\leftarrow$ \labelfont{fashion}\\
			& Negative & & \labelfontinv{sports} $\leftarrow$ \labelfont{quality}, \labelfont{tabloid}\\
			\hline\hline
			\rule{0pt}{1.1em} 
			\multirow{3}{*}{Multi-label} & Partial & \multirow{3}{*}{Label-independent} & \labelfont{quality}, \labelfont{fashion}    $\leftarrow$     University, Female  \\
			& \multirow{2}{*}{Complete} & & \labelfont{quality}, \labelfontinv{tabloid}, \labelfont{fashion}, \labelfontinv{sports}    $\leftarrow$    \\\noalign{\vskip -0.05em}
& & & \hspace*{\fill}University, Female  \\
			\hline \rule{0pt}{1.1em} 
			\multirow{2}{*}{Multi-label} & \multirow{2}{*}{Partial} & Partially Label-dependent & \labelfont{tabloid}, \labelfontinv{sports} $\leftarrow$ \labelfont{fashion}, Children \\
			&  & Fully Label-dependent & \labelfontinv{fashion}, \labelfontinv{sports} $\leftarrow$ \labelfontinv{quality}, \labelfontinv{tabloid} \\ 
			\hline
		\end{tabular}
	}
\end{table}

Table~\ref{tab:rule_forms} shows examples of different types of rules that can be learned from the sample dataset shown in Figure~\ref{tab:ML-example}.
One can distinguish rules according to several dimensions:
\begin{itemize}[noitemsep]
	\item[--] \emph{multi-label} vs.\ \emph{single-label}: Does the head of the rule contain only a single or multiple predictions?
	\item[--] \emph{positive} vs.\ \emph{negative}: Can we predict only the presence of labels or also their absence?
	\item[--] \emph{dependent} vs.\ \emph{independent}: Do the predictions in part or fully depend on other labels? 
\end{itemize}

The predictions in the head of the rule may also have different semantics. We differentiate between \emph{full predictions} and \emph{partial predictions}.
\begin{itemize}
	\item \textit{full predictions:} Each rule predicts a full label vector $\hat{\catsiv}$, i.e., if a label $\cat_{i}$ is not contained in the head, its absence is predicted, i.e.,  $\catv_{i} = 0$.
	\item \textit{partial predictions:} Each rule predicts the presence or absence of the label only for a subset of the possible labels. For the remaining labels the rule does not make a prediction (but other rules might).
\end{itemize}
For denoting absence of labels, we will sometimes also use a bar above the labels, i.e., $\overline{\cat}$ denotes that label $\cat$ is predicted as non-relevant or not observed. 
We also allow $\catv=\;?$ in heads $\catsiv$ and $\catsi \subset \cats \ \cup \ \{\overline{\cat_1},\ldots,\overline{\cat_\ncats}\}$ to denote that certain labels are not concerned by a rule, i.e., that the label is neither predicted as present nor as absent.

%
Alternative categorizations of dependencies are possible. For example,
	\citet{jf:PL-08-WS-Park} categorized full label dependencies into \emph{subset constraints} $\cati_i \leftarrow
	\cati_j$ (the instances labeled with $\cati_j$ are a subset of
	those labeled with $\cati_i$) and \emph{exclusion constraints} $\overline{\cati_i} \leftarrow
	\cati_j$ (the  instances labeled with $\cati_i$ are
	disjoint from
	those labeled with $\cati_j$), which can be readily expressed in a rule-based manner. Fully label dependencies are also known as \emph{global
		dependencies}
	whereas partially label-dependent rules are also known as \emph{local} and \emph{semi-local dependencies}.
    For example, in
	rule~(\ref{eq:rule}), the features used in
	the body of the rule $\fea^{(1)}, \fea^{(2)},  \dots , \fea^{(i)}$ form the local context in which the dependency $\cati^{(i+1)},  \dots , \cati^{(i+j)}
	\rightarrow
	\cati^{(i+j+1)}, \cati^{(i+j+2)},  \dots , \cati^{(k)}$ holds.
	

\subsection{Challenges for Multi-Label Rule Learning}
\label{sec:challenges}

Proposing algorithms that directly learn sets of such rules is a very challenging problem, which involves several subproblems that are not or only inadequately addressed by existing rule learning algorithms.

Firstly, rule-based models expand the class of
dependency models that are commonly used when learning from
multi-label data. As already explained, one commonly distinguishes between
\emph{conditional} and \emph{unconditional} label dependencies
\owncite{dembczynski12PCCdependence}, where
the former is of a \emph{global} nature and holds (unconditionally) in the entire instance space (regardless of any features of the instances), whereas the latter is of a \emph{local} nature and only holds for a specific instance.
By modeling semi-local dependencies that hold in a certain part of the instance space, i.e., for subgroups of the population characterized by specific features, rule-based models allow for a smooth interpolation between these two extremes.
Such dependencies can be formulated elegantly via rules that mix regular features and labels in the condition part of the rule, as illustrated in Table~\ref{tab:rule_forms}.
Besides, rule models offer interesting alternatives for the interpretation of dependencies. While the conventional definition of dependency is based on
probabilistic concepts, rule models are typically associated with
deterministic dependencies. Yet, single rules may also be equipped with  probabilistic semantics (e.g., the condition specified in the head of the rule holds with a certain probability within the region specified in the rule body).

Secondly, a rule-based formulation adds a considerable amount of
flexibility to the learning process. Contrary to single-label
classification, there is a large variety of loss
	functions according to which the performance of multi-label learning
algorithms can be assessed (see Section~\ref{sec:evaluation}).
In a rule-based framework, a loss-minimizing head
could be found for individual rules, so that the same rule body could
be adapted to different target loss functions. 
Conversely, while conventional rule learning heuristics are targeted
towards minimizing classification error aka 0/1-loss, their adaptation
to different multi-target loss functions is not straightforward. Moreover,
different loss functions may require different heuristics
in the underlying rule learner. 

Moving from learning single rules, a process which is also known
as subgroup discovery, to the learning of rule sets adds
another layer of complexity to the rule learning algorithms
\citep{jf:Dagstuhl-04}.
Even an adaptation of the simple and
straightforward covering strategy, which is predominantly used for
learning rule sets in inductive rule learning \citep{jf:AI-Review}, is a
non-trivial task. 
For example, when learning rules
with partial label heads, one has to devise
strategies for dealing with examples that are partially covered, in
the sense that some of their labels are covered by a rule whereas
others are not. One also has to deal with possible
conflicts that may arise from mixed positive and negative rules. Last but not least, one has to recognize and avoid circular dependency
structures, where, e.g., the prediction of label $\cati_i$ depends
on the knowledge of a different label $\cati_j$, which in turn depends on knowledge of $\cati_i$.
Algorithmically, we consider this the most challenging problem.

Finally, rule-based representations are directly interpretable and
comprehensible to humans, at least in principle.
Hence, one is able to analyze
the induced rule models, including dependencies between labels discovered in the data, and may greatly benefit from the insight they provide. This is in contrast to many other types of models, for which the key information is not directly accessible.
Interestingly, the possibility to inspect the algorithmic decision-making process and the right for explanation might play a major role in the up-coming European legislation \citep{goodman16rightToExplain}, which might even raise liability issues for manufacturers, owners and users of artificial intelligence systems.\footnote{ \emph{General Data Protection Regulation} 2016/679 and \emph{Civil law rules on robotics} under the ID of 2015/2103(INL).}

We note in passing, however, that while rules are commonly perceived to be more comprehensible than other types of hypothesis spaces that are commonly used in machine learning, the topic of learning \emph{interpretable} rules is still not very well explored \citep{ComprehensibleModels}. For example, in many studies, the comprehensibility of learned rules is assumed to be negatively correlated with their complexity, a point of view that has been questioned in recent work \citep{User-ModelUnderstandability,jf:DS-16-ShortRules,jf:CognitiveBias-Interpretability}. In this chapter, our main focus is on arguing that the fact that input data and labels can be used to formulate explicit mixed-dependency rules has a strong potential for increasing the interpretability of multi-label learning.


\section{Discovery of Multi-Label Rules}
\label{sec:discovery}

In this section, we review work on the problem of discovering individual multi-label rules. In Section~\ref{sec:ML-associations}, we discuss algorithms that are based on association rule discovery, which
allow to quickly find mixed dependency rules. However, for these algorithms it often remains unclear what loss is minimized by their predictions. Other approaches, which we will discuss in Section~\ref{sec:loss-minimization}, aim at discovering loss-minimizing rules.

\subsection{Association Rule-Based Algorithms}
\label{sec:ML-associations}

A simple, heuristic way of discovering multi-label rules is to convert the problem into an association rule discovery problem (cf. Section~\ref{sec:descriptive-rules}).
To this end, one can use the union of labels and features as the basic itemset,
discover all frequent itemsets, and derive association rules from these frequent
itemsets, as most association rule discovery algorithms do. The only modification is that only rules with labels in the head are allowed, whereas potential rules with features in the head will be disregarded.

For instance, \citet{MMAC} and similarly \citet{MLCwARs} induce single-label association rules, based on algorithms for class association rule discovery \citep{CBA,CBA2}.
Their idea is to
use a multiclass, multi-label associative classification approach
where
single-label class association rules are merged to create
multi-label rules.
\citet{RuleBasedMLCwLCM} employ a more complex approach based on an genetic search algorithm which integrates the discovery of multi-label heads into the evolutionary process.
Similarly, \citet{EvoMLCuARM} and \citet{MLCRules}, use evolutionary
algorithms or classifier systems
for evolving multi-label classification rules  thereby avoiding the
problem of devising search algorithms that are targeted towards that
problem.

An associate multi-label rule learner with several possible labels in the head of the rules was developed by \citet{thabtah06MLassociative}. These labels are found in the whole training set, while the multi-label lazy associative approach of \citet{LazyMLC} generates the rules from the neighborhood of a test instance during prediction. The advantage then is that fewer training instances are used to compute the coverage statistics which is beneficial when small rules are a problem as they are often predicted wrong due to whole training set statistics.

A few algorithms focus on discovering global dependencies, i.e., fully label-dependent rules.
\citet{jf:PL-08-WS-Park} use an association rule miner (Apriori) to discover
pairwise subset (implication) and exclusion constraints, which may be viewed as global dependencies.
These are then applied in the classification process to correct predicted label rankings that violate the globally found constraints.
Similarly, \citet{LI-MLC} infer global dependencies between
the labels in the form of association rules, and use them as a
post-processor for refining the predictions of conventional multi-label
learning systems.
\citet{papagiannopoulou15deterministicrelations} propose a method for discovering deterministic positive entailment (implication) and exclusion relationships between labels and sets of labels.

However, while all these approaches allow to quickly discover multi-label rules,
 it remains mostly unclear what multi-label loss the discovered rules are actually minimizing.

\subsection{Choosing Loss-Minimizing Rule Heads}
\label{sec:loss-minimization}

A key advantage of rule-based methods is that learned rules can be flexibly
adapted to different loss functions by choosing an appropriate head for a given rule body.
Partial prediction rules, which do not predict the entire label vector, require particular attention.
Very much in the same way as completeness in terms of covered
examples is only important for complete rule-based theories and not so much for individual rules,
completeness in terms of predicted labels is less of an issue when learning individual rules.
Instead of evaluating the rule candidates with respect to only one target label,
multi-label rule learning algorithms need to
evaluate candidates w.r.t. all possible target variables and choose the best possible head for each candidate.

Algorithmically, the key problem is therefore to find the empirical loss minimizer of a rule, i.e., the prediction that minimizes the loss on the covered examples, i.e.,
we need to find the multi-label head $\catsiv$ which reaches the best possible performance
\begin{equation}
\heur_{max} = \max_{\catsiv} \ \heur(\rul) = \max_{\catsiv} \ \heur(\catsiv \leftarrow B)
\end{equation}
given an evaluation function $\heur(.)$ and a  body $B$.
As recently shown for the case of the F-measure, this problem is highly non-trivial for certain loss functions \owncite{waegeman14optimalF1}. \citet{MCTS-Multilabel}
adapt an algorithm for subgroup discovery so that it can find the top-k multi-label rules, but the quality measure they use is based on subgroup discovery and not related to commonly used multi-label classification losses.

%
%
\label{sec:relationAR}
To illustrate the difference between measures used in association rule discovery and in multi-label rule learning, assume that the rule $\cat_1, \cat_2 \leftarrow B$ covers three examples $(\doc_1,\{\cat_2\})$, $(\doc_2,\{\cat_1,\cat_2\})$ and $(\doc_3,\{\cat_1\})$. In conventional association rule discovery the head is considered to be satisfied for one of the three covered examples ($\doc_2$), yielding a precision/confidence value of $\frac{1}{3}$. This essentially corresponds to subset accuracy. On the other hand, micro-averaged precision would correspond to the fraction of $4$ correctly predicted labels among $6$ predictions, yielding a value of $\frac{2}{3}$.


\subsubsection{Anti-Monotonicity and Decomposability}
\citet{MLC:Rules-AntiMonotonicity} investigate the behavior of multi-label loss functions w.r.t. two such properties, namely anti-monotonicity and decomposability.
The first property which can be exploited for pruning searches---while still being able to find the best solution---is \emph{anti-monotonicity}.
This property is already well known from association rule learning \citep{APriori,FrequentPatternMining-Survey,AssociationRules-Overview} and subgroup discovery \citep{SupervisedDescriptiveRuleDiscovery,SubgroupDiscovery-Survey}.
In the multi-label  context  it basically states that, if we use an anti-monotonic heuristic $\heur$ for evaluating rules, using  adding additional labels to a head cannot improve its value if adding the previous label already decreased the heuristic value.
An even stronger criterion for pruning the searches can be found particularly for decomposable multi-label evaluation measures.
In few words,
\emph{decomposability}
allows to find the best head by combining the single-label heads which reach the equal maximum heuristic value for a given body and set of examples.
Hence, finding the best head for a decomposable heuristic comes at linear costs, as the best
possible head can be deduced from considering each available label separately.

\begin{table}[t!p]
	\label{tab:properties_metrics}
		\caption{Anti-monotonicity and decomposability of selected evaluation functions with respect to different averaging and evaluation strategies.
		}
	\centering
	\begin{tabular}{|c|c|c|c|c|c}
		\cline{1-5}
		\pbox[c][30pt][c]{\textwidth}{\textbf{Evaluation}\\ \centering \textbf{Function}} & \pbox[c][30pt][c]{\textwidth}{\textbf{Evaluation}\\ \centering \textbf{Strategy}} & \pbox[c][30pt][c]{\textwidth}{\textbf{Averaging} \\ \centering \textbf{Strategy}} & \pbox[c][30pt][c]{0.28\textwidth}{\textbf{Anti-}\\ \centering \textbf{Monotonicity}} &
  	\pbox[c][30pt][c]{0.28\textwidth}{\textbf{Decom-}\\ \centering \textbf{posability}} & \\ \hline \hline
		
		\multirow{8}{*}{Precision}        & \multirow{4}{*}{\pbox[c][30pt][c]{\textwidth}{Partial\\ \centering Predictions}}   & Micro-averaging                          & Yes & Yes & \\ 
		&                                   & Label-based                    & Yes & Yes & \\
		&                                   & Example-based                  & Yes & Yes & \\
		&                                   & Macro-averaging                          & Yes & Yes & \\
		\cline{2-5}
		& \multirow{4}{*}{\pbox[c][30pt][c]{\textwidth}{Full\\ \centering Predictions}} & Micro-averaging                          & Yes & ---   & \\
		&                                   & Label-based                    & Yes & ---   & \\
		&                                   & Example-based                  & Yes & ---   & \\
		&                                   & Macro-averaging                          & Yes & ---   & \\
		\cline{1-5}
		\multirow{8}{*}{Recall}           & \multirow{4}{*}{\pbox[c][30pt][c]{\textwidth}{Partial\\ \centering Predictions}}   & Micro-averaging                          & Yes & Yes &   \\
		&                                   & Label-based                    & Yes & Yes &   \\
		&                                   & Example-based                  & --- & ---     &   \\
		&                                   & Macro-averaging                          & Yes & Yes & \\
		\cline{2-5}
		& \multirow{4}{*}{\pbox[c][30pt][c]{\textwidth}{Full\\ \centering Predictions}} & Micro-averaging                          & Yes & ---   &   \\
		&                                   & Label-based                    & Yes & ---   &   \\
		&                                   & Example-based                  & --- & ---     &   \\
		&                                   & Macro-averaging                          & Yes & ---   & \\
		\cline{1-5}
		\multirow{8}{*}{\pbox[c][30pt][c]{\textwidth}{Hamming\\ \centering Accuracy}} & \multirow{4}{*}{\pbox[c][30pt][c]{\textwidth}{Partial\\ \centering Predictions}}   & Micro-averaging                          & Yes & Yes & \\ 
		&                                   & Label-based                    & Yes & Yes & \\
		&                                   & Example-based                  & Yes & Yes & \\
		&                                   & Macro-averaging                          & Yes & Yes & \\
		\cline{2-5}
		& \multirow{4}{*}{\pbox[c][30pt][c]{\textwidth}{Full\\ \centering Predictions}} & Micro-averaging                          & Yes & ---   & \\
		&                                   & Label-based                    & Yes & ---   & \\
		&                                   & Example-based                  & Yes & ---   & \\
		&                                   & Macro-averaging                          & Yes & ---   & \\
		\cline{1-5}
		\multirow{8}{*}{F-Measure}        & \multirow{4}{*}{\pbox[c][30pt][c]{\textwidth}{Partial\\ \centering Predictions}}   & Micro-averaging                          & Yes & Yes &   \\
		&                                   & Label-based                    & Yes & Yes &   \\
		&                                   & Example-based                  & Yes & Yes &   \\
		&                                   & Macro-averaging                          & Yes & Yes & \\
		\cline{2-5}
		& \multirow{4}{*}{\pbox[c][30pt][c]{\textwidth}{Full\\ \centering Predictions}} & Micro-averaging                          & Yes & ---   &   \\
		&                                   & Label-based                    & Yes & ---   &   \\
		&                                   & Example-based                  & Yes & ---   &   \\
		&                                   & Macro-averaging                          & Yes & ---   & \\
		\cline{1-5}
		\multirow{2}{*}{\pbox[c][30pt][c]{\textwidth}{Subset\\ \centering Accuracy}}  & \pbox[c][30pt][c]{\textwidth}{Partial\\ \centering Predictions}                  & \multirow{2}{*}{Example-based} & Yes & ---   &   \\
		\cline{2-2} \cline{4-5}
		& \pbox[c][30pt][c]{\textwidth}{Full\\ \centering Predictions}                 &                                          & --- & ---     &   \\
		\cline{1-5}
	\end{tabular}
\end{table}

Decomposability is a stronger criterion, i.e., an evaluation measure that is decomposable is also anti-monotonic.
Decomposable multi-label evaluation measures include micro-averaged rule-dependent precision, F-measure, and Hamming accuracy.
Subset accuracy only fulfills the anti-monotonicity property.
This can also be seen from Table~\ref{tab:properties_metrics}, which
shows for a large variety of evaluation measures if maximizing them for a given body can benefit from both criteria.
Detailed proofs are provided by \citet{MLC:Rules-AntiMonotonicity} and \citet{ma:rapp}.


\subsubsection{Efficient Generation of Multi-Label Heads}
\label{sec:rulesearch}
To find the best head for a given body different label combinations must be evaluated by calculating a score based on the used averaging and evaluation strategy. The algorithm described in the following performs a breadth-first search by recursively adding additional label attributes to the (initially empty) head and keeps track of the best rated head. Instead of performing an exhaustive search, the search space is pruned according to the findings in Section~\ref{sec:ML-associations}. When pruning according to anti-monotonicity unnecessary evaluations of label combinations are omitted in two ways: On the one hand, if adding a label attribute causes the performance to decrease, the recursion is not continued at deeper levels of the currently searched subtree. On the other hand, the algorithm keeps track of already evaluated or pruned heads and prevents these heads from being evaluated in later iterations. When a decomposable evaluation metric is used no deep searches through the label space must be performed. Instead, all possible single-label heads are evaluated in order to identify those that reach the highest score and merge them into one multi-label head rule.

\newcommand\solidline[1][0.35cm]{\rule[0.5ex]{#1}{.4pt}}
\newcommand\dashedline{\mbox{
		\solidline[0.5mm]\hspace{0.8mm}\solidline[0.5mm]\hspace{0.8mm}\solidline[0.5mm]}}
\tikzstyle{level 1}=[level distance=1.3cm, sibling distance=3.8cm]
\tikzstyle{level 2}=[level distance=1.7cm, sibling distance=2.2cm]
\tikzstyle{level 3}=[level distance=1.9cm, sibling distance=2.1cm]
\tikzstyle{level 4}=[level distance=1.9cm]
\tikzstyle{bag} = [text width=5.0em, text centered]
\tikzstyle{circled-bag} = [bag, circle, dashed, draw=black, inner sep=0pt]
\tikzstyle{arrow} = [-{Latex[scale=1.5]}]
\begin{figure}[th]
	\centering
	\resizebox{\textwidth}{!}{
		\begin{tikzpicture}
		\node[bag] {$\boldsymbol{\emptyset}$}
		child[black,growth parent anchor={west}] {
			node(a2)[bag,black] {$\boldsymbol{\{\cat_{1}\}}$ \\ $h=\frac{2}{3}$}
			child[black] {
				node[circled-bag,black] {$\boldsymbol{\{\cat_{1},\cat_{2}\}}$ \\ $h=\frac{2}{3}$}
				child[red,growth parent anchor={center}] {
					node(a)[bag,black] {$\boldsymbol{\{\cat_{1},\cat_{2},\cat_{3}\}}$ \\ $h=\frac{5}{9}$}
					child[red] {
						node[bag,black] {$\boldsymbol{\{\cat_{1},\cat_{2},\cat_{3},\cat_{4}\}}$ \\ $h=\frac{5}{12}$}
						edge from parent[arrow]
					}
					edge from parent[arrow]
				}
				child[red] {
					node(b)[bag,black] {$\boldsymbol{\{\cat_{1},\cat_{2},\cat_{4}\}}$ \\ $h=\frac{4}{9}$}
					edge from parent[arrow]
				}
				edge from parent[arrow]
			}
			child[red,growth parent anchor={center}] {
				node(c)[bag,black] {$\boldsymbol{\{\cat_{1},\cat_{3}\}}$ \\ $h=\frac{1}{2}$}
				child[red] {
					node[bag,black] {$\boldsymbol{\{\cat_{1},\cat_{3},\cat_{4}\}}$ \\ $h=\frac{1}{3}$}
					edge from parent[arrow]
				}
				edge from parent[arrow]
			}
			child[red] {
				node[bag,black] {$\boldsymbol{\{\cat_{1},\cat_{4}\}}$ \\ $h=\frac{1}{3}$}
				edge from parent[arrow]
			}
			edge from parent[arrow]
		}
		child[black] {
			node[bag,black] {$\boldsymbol{\{\cat_{2}\}}$ \\ $h=\frac{2}{3}$}
			child[red] {
				node[bag,black] {$\boldsymbol{\{\cat_{2},\cat_{3}\}}$ \\ $h=\frac{1}{2}$}
				child[red] {
					node[bag,black] {$\boldsymbol{\{\cat_{2},\cat_{3},\cat_{4}\}}$ \\ $h=\frac{1}{3}$}
					edge from parent[arrow]
				}
				edge from parent[arrow]
			}
			child[red] {
				node(d)[bag,black] {$\boldsymbol{\{\cat_{2},\cat_{4}\}}$ \\ $h=\frac{1}{3}$}
				edge from parent[arrow]
			}
			edge from parent[arrow]
		}
		child[black,xshift=-1.0cm] {
			node[bag,black] {$\boldsymbol{\{\cat_{3}\}}$ \\ $h=\frac{1}{3}$}
			child[red] {
				node(e)[bag,black] {$\boldsymbol{\{\cat_{3},\cat_{4}\}}$ \\ $h=\frac{1}{6}$}
				edge from parent[arrow]
			}
			edge from parent[arrow]
		}
		child[black,xshift=-2.9cm,yshift=+0.06cm] {
			node(b2)[bag,black,yshift=+0.0cm] {$\boldsymbol{\{\cat_{4}\}}$ \\ $h=0$}
			edge from parent[arrow]
		};
		\draw[dashed] ([yshift=-0.5cm] a.south west) to ([xshift=0.6cm, yshift=-0.5cm] b.south) to[out=0,in=180] ([xshift=-0.5cm, yshift=-0.5cm] c.south) to ([yshift=-0.5cm] d.south) to ([xshift=1.5cm, yshift=-0.5cm] e.south east);
		\draw[solid] ([xshift=-1.0cm, yshift=-0.4cm] a2.south west) to ([xshift=+1.5cm, yshift=-0.4cm] b2.south west |-  a2.south east);
		
		\node(table) at ([xshift=-0.4cm,yshift=-2.3cm] e.south east) {
			\begin{tabular}{c c|c c c c|}
			\cline{3-6}
			& & $\catv_{1}$ & $\catv_{2}$ & $\catv_{3}$ & $\catv_{4}$ \\
			\hline
			\multicolumn{1}{|c|}{\multirow{3}{*}{Not covered}} & $\catsiv_{1}$ & 0 & 1 & 1 & 0 \\
			\multicolumn{1}{|c|}{} & $\catsiv_{2}$ & 1 & 1 & 1 & 1 \\
			\multicolumn{1}{|c|}{} & $\catsiv_{3}$ & 0 & 0 & 1 & 0 \\
			\hline
			\multicolumn{1}{|c|}{\multirow{3}{*}{Covered}} & $\catsiv_{4}$ & 0 & 1 & 1 & 0 \\
			\multicolumn{1}{|c|}{} & $\catsiv_{5}$ & 1 & 1 & 0 & 0 \\
			\multicolumn{1}{|c|}{} & $\catsiv_{6}$ & 1 & 0 & 0 & 0 \\
			\hline
			\end{tabular}
		};
		\end{tikzpicture}
	}
	\caption{Search through the label space
		$2^\cats$ with  $\cats= \{\cat_{1},\cat_{2},\cat_{3},\cat_{4}\}$
		using micro-averaged precision of partial predictions. The examples corresponding to label sets $\catsiv_{4},\catsiv_{5},\catsiv_{6}$ are assumed to be covered, whereas those of $\catsiv_{1},\catsiv_{2},\catsiv_{3}$ are not. The dashed line (\protect\dashedline) indicates label combinations that can be pruned with anti-monotonicity, the solid line (\protect\solidline) corresponds to decomposability.}
	\label{figure_example}
\end{figure}
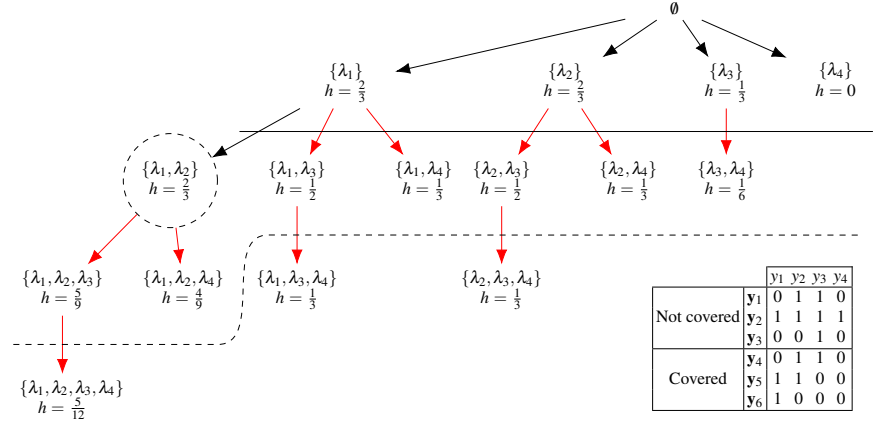

Figure~\ref{figure_example} illustrates how the algorithm prunes a search through the label space using anti-monotonicity and decomposability. The nodes of the given search tree correspond to the evaluations of label combinations, resulting in heuristic values $\heur$. The edges correspond to adding an additional label to the head which is represented by the preceding node. As equivalent heads must not be evaluated multiple times, the tree is unbalanced.


\section{Learning Predictive Rule-Based Multi-label Models}
\label{sub:stacking}

Predictive, rule-based theories are formed by combining individual
rules into a theory. Such an aggregation step is necessary because each individual
rule will only cover a part of the example space.
When mixed dependency rules, i.e., rules with both labels and
features in the rule bodies, are combined into a predictive theory,
several problems arise that make the problem considerably harder than
the aggregation of local rules into a global rule-based model \citep{jf:Dagstuhl-04}.

As a very simple example, consider the case when two labels
$\cati_{i}$ and $\cat_{j}$ always co-occur in the training data. The
algorithms discussed in the previous section would then find the inclusion constraints
$\cat_{i} \rightarrow \cat_{j}$ and $\cat_{j} \rightarrow \cat_{i}$.
These are valid and interesting insights into the domain, but in a
predictive setting, they will not help to identify both labels as positive unless at least one of the two can be predicted by another rule.\footnote{Similar problems have been encountered in inductive logic
	programming, where the learning of recursive and multi-predicate programs has received some attention
	\citep{RecursiveILP,
		MPL,
		Foil-Recursion}.}

As shown by this example, the problem of circular reasoning is a major concern in the inference with mixed dependency rules. There are two principal ways for tackling this problem.
	The simplest strategy is
	to avoid circular dependencies from the very beginning. This means that rules discovered in the learning process have to be organized in a structure that prevents cycles or, alternatively, that additional rules have to be
	learned with certain constraints on the set of valid conditions in the
	rule body.

	Another way of tackling this problem is to allow for circular dependencies and generalize the inference strategy in a suitable manner.
This approach has not yet received much attention in the literature. One notable exception is the work of \citet{montanes14DBR} who realized this idea in so-called dependent binary relevance (DBR) learning, which is based on techniques similar to those used in conditional dependency networks \citep{MLCDN}.

In this section, we will describe two different approaches for tackling the first problem.
One, which we call \emph{layered learning} tries to avoid label cycles by requiring an initial guess for labels regardless of any label dependence (Section~\ref{sec:layered}).\com{ich musste das argument anpassen (vorher, imposing an order on the learned rules). moeglicherweise passen nun auch nicht die referenzen zu DBR und CNN bzw. generell der hinweis auf circular dependencies.}
While this approach is rather coarse in that batches of rules are learned, we will then also consider approaches that try to adapt the covering or separate-and-conquer strategy, which is frequently used in inductive rule learning (Section~\ref{sec:mlcseco}).


\subsection{Layered Multi-label Learning}
\label{sec:layered}

The recently very popular classifier chains \citep[CC;][]{read11classifierchains} were found to be an effective approach for avoiding label cycles. Their key idea is to
use an arbitrary order
$\cat_{1} , \cat_{2} , \dots , \cat_{\ncats}$ on the labels, and
learn rules that involve predictions for $\cat_{i}$ only from the
input attributes and all labels $\cat_{j}, j < i$. This, however, has
some obvious disadvantages, which have been addressed by many variants that have been investigated in the literature.

One drawback is the (typically randomly chosen) predetermined, fixed order of the classifiers (and hence the labels) in the chain, which makes it impossible to learn dependencies in the opposite direction.
This was already recognized by \citet{LabelDepRuleLearning},
who built up a very similar system in order to learn multiple dependent concepts. In this case, the chain on the labels was determined beforehand by a statistical analysis of the label dependencies.
Still, using a rule learner for solving the resulting binary problems would only allow to induce rules between two labels in one direction.


\subsubsection{Stacked Binary Relevance}
An alternative approach without this limitation is to use two levels of classifiers: the first one tries to predict labels independently of each other, whereas the second level of classifiers makes additional use of the predictions of the previous level.
More specifically, the training instances for the second level are expanded by the label information of the other labels, i.e., a training example $\doc$ for label $\catv[i]$ is transformed into
$(\doci[1], \ldots, \catv[1],\ldots,\catv[i-1],\catv[i+1],\ldots,\catv[\ncats])$.
During training, the prediction of the first level of classifiers is used as additional features for the second level, i.e., the final prediction
$\hat{\catv}_{j}$ depends on predictions $\classifier_i(\doc)$ and $\classifier'_j(\doc,\classifier_1(\doc),\ldots,\classifier_\ncats(\doc))$.
Hence, each label will be characterized by two sets of rule models, namely the rules $\sRul_i$ which depend only on instance
features, and a second set of rule models $\sRul^*_i$ depending (possibly) also on other
labels. $\sRul_i$ can then provide the predictions that are necessary for executing the rules in $\sRul^*_i$. \citet{loza14MLRL,ML:Rules-Stacking} refer to this technique as \emph{stacked binary relevance} (SBR) in contrast to plain, unstacked binary relevance learning.

\begin{figure}[t!]
\begin{minipage}[c]{\textwidth}
\begin{minipage}[t]{.49\textwidth}
\fullrule
\begin{tabular}{l}
\labelfont{quality}    $\leftarrow$    University  \\
\labelfontinv{quality}    $\leftarrow$    Single  \\
\labelfont{quality}    $\leftarrow$    Female  \\
\labelfont{quality}    $\leftarrow$    Secondary  \\
\labelfontinv{quality}   $\leftarrow$   true  \\
---\\
\labelfont{quality}    $\leftarrow$    University  \\
\labelfont{quality}    $\leftarrow$    \labelfont{sports}, Secondary  \\
\labelfontinv{quality}    $\leftarrow$    Male  \\
\labelfontinv{quality}   $\leftarrow$   true  \\
\end{tabular}
\fullrule
\end{minipage}%
\hspace*{\fill}
\begin{minipage}[t]{.49\textwidth}
\fullrule
\begin{tabular}{l}
\labelfontinv{tabloid}    $\leftarrow$    Primary, Single  \\
\labelfontinv{tabloid}    $\leftarrow$    Female, Married  \\
\labelfontinv{tabloid}    $\leftarrow$    Divorced, Secondary, $\overline{\text{Children}}$  \\
\labelfont{tabloid}   $\leftarrow$   true  \\
---\\
\labelfontinv{tabloid}    $\leftarrow$    Primary, Single  \\
\labelfont{tabloid}    $\leftarrow$    \labelfontinv{quality}   $\leftarrow$   true  \\
\labelfontinv{tabloid}    $\leftarrow$    Female, Married  \\
\labelfontinv{tabloid}    $\leftarrow$    Secondary, Divorced  \\
\labelfont{tabloid}   $\leftarrow$   true  \\
\end{tabular}
\fullrule
\end{minipage}%
\newline

\begin{minipage}[t]{.49\textwidth}
\fullrule
\begin{tabular}{l}
\labelfontinv{fashion}    $\leftarrow$    Male  \\
\labelfontinv{fashion}    $\leftarrow$    $\overline{\text{Children}}$   \\
\labelfont{fashion}   $\leftarrow$   true  \\
---\\
\labelfontinv{fashion}    $\leftarrow$    Male  \\
\labelfontinv{fashion}    $\leftarrow$    $\overline{\text{Children}}$, \labelfontinv{tabloid}   $\leftarrow$    \\
\labelfont{fashion}   $\leftarrow$   true  \\
\ \\
\end{tabular}
\fullrule
\end{minipage}%
\hspace*{\fill}
\begin{minipage}[t]{.49\textwidth}
\fullrule
\begin{tabular}{l}
\labelfont{sports}    $\leftarrow$    Divorced, Secondary, $\overline{\text{Children}}$  \\
\labelfont{sports}    $\leftarrow$    Children, Male  \\
\labelfontinv{sports}   $\leftarrow$   true  \\
---\\
\labelfontinv{sports}    $\leftarrow$    \labelfontinv{quality}, \labelfont{tabloid}    \\
\labelfont{sports}    $\leftarrow$    Secondary  \\
\labelfont{sports}    $\leftarrow$    Children , Male  \\
\labelfontinv{sports}   $\leftarrow$   true  \\
\end{tabular}
\fullrule
\end{minipage}%
\end{minipage}%
	\caption{Rule set obtained from layered learning on the example dataset (Figure~\ref{figure_example}). Decision lists from first and second level are separated by ---.}
	\label{fig:SBR}
\end{figure}

Figure~\ref{fig:SBR} shows a rule set that can be obtained with SBR for the sample dataset of Figure~\ref{figure_example}. One can see that two separate decision lists are learned for each label using a conventional rule learner such as \ripper\ \cite{Ripper}. The top list $\sRul_i$ is learned only from the input features, and the bottom part $\sRul^*_i$ is learned from input features and the predictions originating from the top rule set.

Despite being able to learn, in contrast to CC, relationships in either direction and in any constellation, this method still has its shortcomings.
Firstly, it requires comprehensive predictive rules for each label on the first level even though the labels may effectively be predicted based on other labels.
For example, assume the global relation $\cat_{i} \leftarrow \cat_{j}$, the approach would need to learn a rule model $\sRul_j$ once for predicting $\cat_j$ and once implicitly as part of $\sRul_i$.

Secondly,
a limitation of the stacking approach may appear when circular dependencies exist between labels.
A very simple  example is if two labels exclude each other, i.e., if
both relationships $\cat_i \leftarrow \overline{\cat_j}$, and  $\cat_j \leftarrow \overline{\cat_i}$ hold.\com{actually, the notation is not complete if you see that as rule. we would have to add $\overline{\cat_i} \leftarrow \cat_j$ and $\overline{\cat_j} \leftarrow \cat_i$. but I propose to leave it like this at the moment.}
Such rules could lead to contradictions and inconsistent states.
For instance, assume that both labels $\cat_i$ and $\cat_j$ were predicted as relevant at the first layer.
Both predictions would be flipped at the next layer, leading again to an (irresolvable) inconsistency   according to the learned rules.

Another problem that needs to be addressed by layered algorithms is
the problem of \emph{error propagation}:
If the prediction for a label $\cat_{j}$ depends on another label $\cat_{i}$, a mistake on the latter is likely to imply a mistake on the former \citep{MLC:CC-ErrorPropagation}.

Finally and most notably, the method of \citep{ML:Rules-Stacking} is limited to produce single-label head rules.

Several variants were proposed in the literature, which in deviation from the basic technique may use predictions instead of the true label information as input \citep{montanes14DBR}, only the predictions in a pure stacking manner \citep{godbole04stacking}, or Gibbs sampling instead of the first level of classifiers \citep{MLCDN}.
\emph{Dependent binary relevance} \citep{montanes14DBR} and \emph{conditional dependency networks} \citep{MLCDN}
are particularly concerned with estimating probability distributions (especially joint distribution). They both use logistic regression
as their base classifier, which is particularly adequate for estimating probabilities.
This type of models are obviously much harder to comprehend than rules, especially for higher number of input features. Therefore, the label dependencies would
remain hidden somewhere in the model, even though they may have been taken
into account and accurate classifiers may have been obtained.
%
A more general approach
is to integrate the stacking of label features directly into the covering loop. Adaptations of the separate-and-conquer strategy to the multi-label case will be discussed in the next section.

\subsection{Multi-Label Separate-and-Conquer}
\label{sec:mlcseco}

The most frequently used strategy for learning a rule-based
predictive theory is the so-called \emph{covering} or \emph{separate-and-conquer}
strategy, which is either integrated into the rule learning algorithm
\citep{jf:AI-Review} or used as a post-processor for selecting a
suitable subset among previously learned rules \citep{CBA}.
Although it may seem quite straightforward, its adaptation to the
multi-label case is only trivial if complete
rules are learned, i.e., if each rule predicts a complete assignment
of the label vector.
In this case, one may learn a decision list with
the covering strategy, which removes all examples that are
covered by previous rules before subsequent rules are learned. In this way,
the learning strategy essentially mirrors the sequential nature in which predictions
are made with a decision list.
In the context of multi-label classification, this strategy corresponds to 	applying the well known \emph{label powerset} transformation which converts each label combination in the data into a meta-label and then solves the resulting multiclass problem \citep[cf.][]{tsoumakas10MLoverview}.

However, in the case of partial-prediction rules, the
situation becomes considerably more complex.  One can, e.g., not simply remove all
examples that are covered by a rule, because the rule will in general
only predict a subset of the relevant labels. An alternative strategy
might be to remove all predicted \emph{labels} from the examples that are covered by the rule: if a rule $\rul: \hat{\catsiv} \leftarrow B$ covers a training example $(\doc,\catsiv)$, the example is not removed but replaced with the example $(\doc,\catsiv \setminus \hat{\catsiv})$. In this way, each training example remains in the training set until all of its labels
are covered by at least one rule.
However, even this strategy may be
problematic, because
removing labels from covered training instances in this way may distort the label dependencies in the training data.

\begin{figure}[tb]
	\hrule
	\smallskip
	\small
	\begin{algorithmic}[1]
		\Require{New training example pairs $\trainset=\{(\doc[1],\catsiv_1),  \ldots ,(\doc[\ndocs], \catsiv_\ndocs)\}$
		}
		\State $\trainset=\{(\doc[1],\hat\catsiv_1),  \ldots ,(\doc[\ndocs], \hat\catsiv_\ndocs)\}$ with $\hat\catsiv_i=(?,?,\ldots,?)$, $i=1\ldots\ndocs$
		\While{$\trainset$ not empty} 
		\State $\rul \gets \textit{findBestGlobalRule}(\trainset)$ \Comment find best rule by refining rule body (and head) w.r.t. some heuristic $h$
		\State apply $\rul$: apply header on covered $\doc[i] \in \trainset$ and put them into $\trainset_{cov}$
		\If{enough $\doc[i]$ in $\trainset_{cov}$ with fully covered labels, i.e.,  $\forall j.\ (\hat\catsiv_i)_j\ \neq\ ?$,}
		\State make $\rul$ full prediction rule and do not add $\trainset_{cov}$ to $\trainset$
		\Else
		\State re-add $\trainset_{cov}$ to $\trainset$
		\EndIf
		\State add $\rul$ to decision list $\sRul$
		\EndWhile
		\State \textbf{return} decision list $\sRul$
	\end{algorithmic}
	\hrule
	\caption{General training algorithm for the multi-label separate-and-conquer algorithm.}\label{alg:train}
\end{figure}


By using separate-and-conquer strategy to induce a rule model, two of the shortcomings of the layered approach from the previous section are addressed.
Firstly, the iterative, non-parallel induction of rules in the covering process ensures that redundant rules are avoided because of the separation step.
Secondly, cyclic dependencies cannot longer harm the induction or prediction process since
the order in which labels are covered or predicted 
is naturally chosen by the covering process. Similarly, the learner may also dynamically model local label dependencies and does not depend on a global order as in classifier chains.

\subsubsection{A Multi-Label Covering Algorithm}
	

Figure~\ref{alg:train} shows the multi-label covering algorithm
proposed by \citet{ML:Rules-Stacking}. The algorithm essentially proceeds as sketched described above, i.e., covered examples are not removed entirely but only the subset of predicted labels is deleted from the example.

For learning a new multi-label rule (line 3), the algorithm performs a top-down greedy search, starting with the most general rule. By adding conditions to the rule's body it can successively be specialized, resulting in fewer examples being covered. Potential conditions result from the values of nominal attributes or from averaging two adjacent values of the sorted examples in case of numerical attributes. Whenever a new condition is added, a corresponding single- or multi-label head that predicts the labels of the covered examples as accurate as possible must be found (cf. Section~\ref{sec:loss-minimization} and, in particular, Figure~\ref{figure_example}).

%
%
%
If a new rule is found, the predicted labels from the examples are marked as covered by these rules, i.e., $(\catsiv_i)_j$ are set to 0 or 1, respectively.
As
depicted in lines 5--6 in the pseudo-code of Figure~\ref{alg:train},
only examples for which enough labels have been predicted can be entirely removed from the training set. A rule that predicts many of such examples is marked as full prediction rule, which means that the execution of the decision list may stop after this rule has fired.

%
%

\begin{figure}[t!]
	\centering

	\subfigure[Single-label head rules (read column-wise)\label{fig:dl-single}]{%
		\begin{minipage}[t]{.5\textwidth}
\fullrule
			\begin{tabular}{l}
				\labelfontinv{fashion}    $\leftarrow$    Male  \\
				\labelfontinv{sports}    $\leftarrow$    Children \\
				\labelfontinv{quality}    $\leftarrow$    Primary  \\
				\labelfont{quality}    $\leftarrow$    University  \\
				\labelfontinv{sports}    $\leftarrow$    \labelfontinv{quality}   $\leftarrow$   true  \\
				\labelfontinv{tabloid}    $\leftarrow$    Female, $\overline{\text{Children}}$  \\
				\labelfont{fashion}    $\leftarrow$    Children   \\
				\labelfontinv{sports}    $\leftarrow$    University  \\
				\labelfontinv{quality}    $\leftarrow$    Single  \\
				\labelfont{quality}    $\leftarrow$    $\overline{\text{Children}}$  \\
				\labelfont{sports}    $\leftarrow$    Divorced  \\
				\labelfont{tabloid}    $\leftarrow$    Married, Male  \\
			\end{tabular}
\fullrule
		\end{minipage}%
		\begin{minipage}[t]{.5\textwidth}
\fullrule
			\begin{tabular}{l}
				\labelfontinv{tabloid}    $\leftarrow$    Primary  \\
				\labelfont{tabloid}    $\leftarrow$    Single  \\
				\labelfontinv{tabloid}    $\leftarrow$    Secondary  \\
				\labelfontinv{sports}   $\leftarrow$   true  \\
				\labelfont{tabloid}    $\leftarrow$    Divorced  \\
				\labelfontinv{tabloid}   $\leftarrow$   true  \\
				\labelfontinv{fashion}    $\leftarrow$    Married  \\
				\labelfontinv{fashion}, *    $\leftarrow$    \labelfont{sports}  \\
				\labelfont{fashion}   $\leftarrow$   true  \\
				\labelfont{quality}, *    $\leftarrow$    \labelfont{tabloid}   \\
				\labelfontinv{quality}, *   $\leftarrow$   true  \\
				\ \\
			\end{tabular}
\fullrule
		\end{minipage}
	}

	\subfigure[Multi-label head rules \label{fig:dl-multi}]{%
		\begin{minipage}[t]{\textwidth}
\fullrule
			\begin{tabular}{l}
				\labelfontinv{quality}, \labelfontinv{fashion}, \labelfontinv{sports}    $\leftarrow$    Primary  \\
				\labelfont{quality}, \labelfontinv{sports}    $\leftarrow$    University  \\
				\labelfontinv{fashion}    $\leftarrow$    Male  \\
				\labelfontinv{quality}, \labelfont{tabloid}, \labelfont{fashion}, \labelfontinv{sports}    $\leftarrow$    Single, Secondary  \\
				\labelfont{quality}, \labelfontinv{tabloid}, \labelfont{sports}    $\leftarrow$    Female  \\
				\labelfont{quality}, \labelfont{tabloid}, \labelfont{fashion}, \labelfontinv{sports}    $\leftarrow$    Married  \\
				\labelfont{quality}, \labelfontinv{tabloid}, \labelfontinv{fashion}, \labelfont{sports}    $\leftarrow$    Secondary, $\overline{\text{Children}}$  \\
				\labelfontinv{quality}, \labelfont{fashion}, \labelfontinv{sports}   $\leftarrow$   true  \\
				\labelfontinv{tabloid}, *    $\leftarrow$    \labelfontinv{quality}   $\leftarrow$   true  \\
				\labelfont{tabloid}, *   $\leftarrow$   true  \\
			\end{tabular}
\fullrule
		\end{minipage}
	}
	\caption{Decision lists induced from the sample dataset of Figure~\ref{figure_example} with precision as heuristic.
		The stars (*) indicate full prediction rules, after which the prediction stops if the rule fires.}
	\label{fig:dls}
\end{figure}

To classify test examples, the learned rules are applied in the order of their induction. If a rule fires, the labels in its head are applied unless they were already set by a previous rule.\footnote{This corresponds to the default strategy in
classification rule learning, where rules are appended at the end of a list. Note however, that there are also good arguments for prepending rules at the beginning of the list, so that, e.g., exceptions are handled before the general rule \citep{Prepend}.}
The process continues with the next rule in the multi-label decision list 
until either a specially marked full prediction rule is encountered or all rules of the decision list have been processed.

Note that if we had only a single binary (or multiclass) label, i.e. $\ncats=1$, the described algorithm would behave exactly as the original separate-and-conquer approach.
However, for $\ncats>1$ the algorithm re-adds partially and even fully covered examples instead of removing them (line 8).
These examples may serve as an anchor point for subsequent rules and facilitate in such a manner the rule induction process.
Moreover, this step enables the algorithm to induce rules which test for the presence of labels.
These type of rules are of particular interest since they explicitly reveal label dependencies discovered in the dataset.

Figure~\ref{fig:dls} shows the results of applying these algorithms to our sample dataset. The top part shows the rules obtained with the single-label head version of \citet{ML:Rules-Stacking}, whereas the lower part shows  those of the multi-label head extension by \citet{MLC:Rules-AntiMonotonicity}.

\begin{table}[t]
\caption{Statistics of the used datasets: name of the dataset, domain of the input instances, number of instances, number of nominal/binary and numeric features, total number of unique labels, average number of labels per instance (cardinality), average percentage of relevant labels (label density), number of distinct label sets in the data.
}	
\label{tab:datastats}
\centering
\resizebox{1.0\textwidth}{!}{
\begin{tabular}{l|cc|cc|c|ccc}
\hline
\footnotesize
\textbf{Name}  &  \textbf{Domain}  &  \textbf{Instances}  &  \textbf{Nominal}  &   \textbf{Numeric}  &  \textbf{Labels}  &  \textbf{Cardinality}  &  \textbf{Density}  &  \textbf{Distinct}
\\
\hline\hline
\ds{emotions}  &  music  &  593  &  0  &  72  &  6  &  1.869  &  0.311  &  27  \\
\ds{scene}  &  image  &  2407  &  0  &  294  &  6  &  1.074  &  0.179  &  15 \\
\ds{flags}  &  image  & 194    & 9    & 10   &  7  &  3.392 & 0.485 & 54 \\
\ds{yeast}  &  biology  &  2417  &  0  &  103  &  14  &  4.237  &  0.303  &  198 \\
\ds{birds}  &  audio & 645  & 2 & 258 & 19 & 1.014 &  0.053 &  133 \\
\ds{genbase}  &  biology  &  662  &  1186  &  0  &  27  &  1.252  &  0.046  &  32 \\
\ds{medical}  &  text  &  978  &  1449  &  0  &  45  &  1.245  &  0.028  &  94 \\
\ds{enron}  &  text  &  1702  &  1001  &  0  &  53  &  3.378  &  0.064  &  753 \\
\ds{CAL500}  &  music  &  502  &  0  &  68  &  174  &  26.0  &  0.150  &  502 \\
\hline
\end{tabular}
}
\end{table}
\section{Case Studies}
\label{sec:eval}

In this section, we show a few sample result obtained with some of the algorithms described in the previous sections on commonly used benchmark data. Our main focus lies on the inspection and the analysis of the induced rule models, and not so much on their predictive accuracy in comparison to state-of-the-art multi-label classification methods (generally, the predictive performance of rule-based models will be lower). We primarily show some sample rule models, but also discuss
statistics on the revealed dependencies.

We experimented with several datasets from the MULAN repository.\footnote{\url{http://mulan.sf.net/datasets.html}}
Table~\ref{tab:datastats} gives a brief overview of the used datasets, along with characteristics such as the number of instances, the number and nature of the attributes, as well as some characteristics on the distribution of labels.
The datasets are from different domains and have varying properties. Details of the data are given in the analysis when needed.

\subsection{Case Study 1: Single-Label Head Rules}
\label{sec:casestudy1}

In the first case study, we compared several single-head multi-label rule learning algorithms, namely
conventional binary relevance (\learner{BR}), the layered algorithm stacked binary relevance (\learner{SBR}), and a separate-and-conquer learner seeking for rules with only a single label in the head (\learner{Single}).
The rule learner \ripper\ \citep{Ripper} was used for finding the label-specific candidate single-head candidate rules. Among these, the best was selected according to the micro-averaged \mbox{F-measure} \citep{ML:Rules-Stacking}.

In the following, we first take a closer look on the actual rules, comparing them to the rules induced separately for each label and by separate-and-conquer. Subsequently, we
put the focus on visualizing dependencies between labels found by the stacking approach.
We refer to \citet{ML:Rules-Stacking} for extensive statistics and more detailed evaluations.

\begin{figure}[t]
\centering
\resizebox{0.8\textwidth}{!}{
\footnotesize
\begin{tabular}{c|l|l}
\hline
\textbf{Approach} & \multicolumn{2}{c}{\ds{yeast}} \\
\hline\hline
\rule{0pt}{1.1em} 
\learner{BR}
 & \multicolumn{2}{l}{\labelfont{Class4} $\leftarrow$ x23 $>$ 0.08, x49 $<$ -0.09} \\
 & \multicolumn{2}{l}{\labelfont{Class4} $\leftarrow$ x68 $<$ 0.05, x33 $>$ 0.00, x24 $>$ 0.00,   x66 $>$ 0.00, x88 $>$ -0.06 } \\
 & \multicolumn{2}{l}{\labelfont{Class4} $\leftarrow$ x3 $<$ -0.03, x71 $>$ 0.03, x91 $>$ -0.01 } \\
 & \multicolumn{2}{l}{\labelfont{Class4} $\leftarrow$ x68 $<$ 0.03, x83 $>$ -0.00, x44$>$ 0.029, x93 $<$ 0.01 } \\
 & \multicolumn{2}{l}{\labelfont{Class4} $\leftarrow$ x96 $<$ -0.03, x10 $>$ 0.01, x78$<$ -0.07 } \\
\hline
\rule{0pt}{1.1em} 
\learner{SBR}
& \multicolumn{2}{l}{\labelfont{Class4} $\leftarrow$ \labelfont{Class3}, $\overline{\text{\labelfont{Class2}}}$  }\\
 & \multicolumn{2}{l}{\labelfont{Class4} $\leftarrow$ \labelfont{Class5}, $\overline{\text{\labelfont{Class6}}}$ }\\
& \multicolumn{2}{l}{\labelfont{Class4} $\leftarrow$ \labelfont{Class3}, \labelfont{Class1}, x22 $>$ -0.02 }\\
\hline
\rule{0pt}{1.1em} 
\learner{Single}
&    \multicolumn{2}{l}{\labelfont{Class4} $\leftarrow$ \labelfont{Class3}, x91 $>$ -0.02, x50 $<$ -0.02, x68 $<$ 0.03}
\\
&  \multicolumn{2}{l}{\labelfont{Class4} $\leftarrow$ \labelfont{Class3}, x90 $>$ -0.02, x77 $<$ -0.04}
\\
&    \multicolumn{2}{l}{\labelfont{Class4} $\leftarrow$  x60 $<$ -0.03, x57 $<$ -0.07, x19 $>$ -0.01 }
\\
\hline
 &  \multicolumn{1}{c|}{\ds{medical}} & \multicolumn{1}{c}{\ds{enron}}\\
\hline\hline
\rule{0pt}{1.1em} 
\learner{BR}
 &  \labelfont{Cough} $\leftarrow$ ``cough'', $\overline{\text{``lobe''}}$   & \labelfont{Joke} $\leftarrow$ ``mail'', ``fw'', ''didn''\\
  &  \labelfont{Cough} $\leftarrow$ ``cough'', ``atelectasis''  &   \\
 & \labelfont{Cough} $\leftarrow$ ``cough'', ``opacity'' & \\
 & \labelfont{Cough} $\leftarrow$  ``cough'', ``airways'' & \\
  & \labelfont{Cough} $\leftarrow$ ``cough'' , $\overline{\text{``pneumonia''}}$, $\overline{\text{``2''}}$ & \\
 &  \labelfont{Cough} $\leftarrow$ ``coughing'' & \\
 &  \labelfont{Cough} $\leftarrow$ ``cough'', ``early'' & \\
\hline
\rule{0pt}{1.1em} 
\learner{SBR}
 &  \labelfont{Cough} $\leftarrow$ ``cough'' , $\overline{\text{\labelfont{Pneumonia}}}$ ,  & \labelfont{Joke} $\leftarrow$ \labelfont{Personal}, ``day'', ``mail'' \\
& \luecke{Cough } $\overline{\text{\labelfont{Pulmonary\_collapse}}}$ , $\overline{\text{\labelfont{Asthma}}}$ &  \\
 & \labelfont{Cough} $\leftarrow$ ``coughing'' \\
 & \labelfont{Cough} $\leftarrow$ \labelfont{Asthma}, ``mild'' & \\
\hline
\rule{0pt}{1.1em} 
\learner{Single}
& \labelfont{Cough} $\leftarrow$ ``cough'', $\overline{\text{``lobe''}}$, $\overline{\text{``asthma''}}$
& \labelfont{Joke} $\leftarrow$ ``didn'', $\overline{\text{``wednesday''}}$\\
& \labelfont{Cough} $\leftarrow$ ``cough'', ``opacity''
& \labelfont{Joke}  $\leftarrow$  \labelfont{Personal}, ``forwarded'' \\
& \labelfont{Cough} $\leftarrow$ ``cough'', ``atelectasis''
\\
& \labelfont{Cough} $\leftarrow$ ``cough'', ``airways''
\\
& \labelfont{Cough} $\leftarrow$ ``cough'', \labelfont{Fever} & \\
\hline
\end{tabular}
}
\caption{Example rule sets for one exemplary label, respectively,  learned by BR, SBR and separate-and-conquer (single).  }
\label{fig:modelexamples}
\end{figure}

\subsubsection{Exemplary Rule Models}

Examples of learned rule sets are shown in Figure~\ref{fig:modelexamples}.
In the case of \ds{yeast}, we see a
much more compact and
less complex rule set for \emph{Class4} for the layered learner \learner{SBR} than for the
independently learned \learner{BR} classifier.
The rule set also seems more appropriate for a domain expert to understand coherences between proteins (instance features) and protein functions (labels).
The separate-and-conquer model \learner{Single} is less explicit in this sense, but it shows that certainly $\labelfont{Class3}$ is an important class for expressing $\labelfont{Class4}$.\footnote{For convenience, we only show the rules with this label in the head.}

Figure~\ref{fig:modelexamples} also shows the models for the diagnosis \emph{Cough} in the \ds{medical} task. This dataset is concerned with the assignment of international diseases codes (ICD) to real, free-text radiological reports. 
Interestingly, the model found by \learner{SBR} reads very well, and the found relationship seems to be even comprehensible to non-experts. For example, the first rule can be read as
\begin{quotation} \it\noindent
If the patient does not have \emph{Pneumonia}, a \emph{Pulmonary\_collapse} or \emph{Asthma} and ``cough''s or is ``coughing'', he just has a \emph{Cough}. Otherwise, he may also have a ``mild'' \emph{Asthma}, in which case he is also considered to have a \emph{Cough}.
\end{quotation}
The theory learned by \learner{Single} is quite similar to the one learned by simple
\learner{BR}, which shows that the textual rules were considered to be more powerful
than the dependency-based rule. Only at the end, a local dependency is learned:
\emph{Cough} only depends on the word ''cough'' if the label for \emph{Fever} has also been set.

In \ds{enron}, which is concerned with the categorization of emails during the Enron scandal, the learned models are generally less comprehensible.
The observed relation between \emph{Personal} and \emph{Joke} can clearly be explained from the hierarchical structure on the topics.

Regarding the sizes of the models, we found between 50 and 100 rules for \ds{yeast} and \ds{medical}, and between 76 (\emph{BR}) and 340 (\emph{Single}) for \ds{enron}. Note, however, that even for \ds{enron} this results in an average of only 6.4 rules per label for the largest rule model.
Moreover, only a fraction of them are necessary in order to track and comprehend the prediction for a particular test instance. For instance, the report
\begin{quotation} \it\noindent
Clinical history: Cough for one month. \\
Impression: Mild hyperinflation can be seen in viral illness or reactive airway disease. Streaky opacity at the right base is favored to represent atelectasis.
\end{quotation}
in the \ds{medical} dataset was classified by experts as normal \emph{Cough}, as well as by the rule sets in Figure~\ref{fig:modelexamples}.
Furthermore, the rule models allow to identify the relationships found by the algorithm responsible for the prediction and even the training examples responsible for finding such patterns.
This type of inspection may facilitate in a more convenient way than with black box approaches the integration of expert feedback---for instance on the semantic coherence \citep{gabriel14coherentrules} or plausibility of rules---and also an interaction with the user.
For example, \citet{master:beckerle}  explored an interactive rule learning process where learned rules could be directly modified by the user,  thereby causing the learner to re-learn subsequently learned rules.


%
%
%

\subsubsection{Visualization of Dependencies}


Figure~\ref{fig:depfig} shows a graphical representation of the label-dependent rules found by $SBR$ on some of the smaller datasets.
Each graph shows a correlation between label pairs.
Labels are enumerated from 1 to the number of labels, and the corresponding label names are shown at the bottom of the coordinate system. 
Blue boxes in the intersection square between a \emph{row label} and a \emph{column label} depict fully label-dependent rules, green boxes show partially label-dependent rules. A colored box at the top corners indicates a rule of the type $$\textit{row label} \leftarrow \ldots,\textit{column label},\ldots,$$ whereas the bottom corners represent the opposite  $$\textit{column label} \leftarrow\ldots, \textit{row label},\ldots$$ rules.
%
As an example, the blue box in the upper left corner of the square in the second row and fourth column in Figure~\ref{fig:depfig}a (\ds{emotions}) indicates that the algorithm found a rule of the type $$\labelfont{happy-pleased}\leftarrow\ldots,\textrm{quiet-still},\ldots,$$
i.e., that quiet or still audio sample cause (possibly together with other factors) happy or pleased emotions.
 Note, however, that the graphs do not show
whether the head or conditions are positive or negative.

In particular in \ds{scene} (Figure~\ref{fig:depfig}b), we find many local dependencies, which also depend on some instance features. This is reasonable, since the task in this dataset is to predict elements of a scenery image, and although some label combinations may be more likely than others, whether an element is present or not will still 
depend on the content of the picture at hand.
In \ds{yeast} the labels seem to be organized in a special way since we encounter the pattern that a label depends on its preceding and the two succeeding labels.
\ds{enron} has a hierarchical structure on its labels, which can be recognized from the vertical and horizontal patterns originating from parent labels.


\input{depfig}

\subsubsection{Discussion}
In our experiments, a layered learning approach such as \learner{SBR} proved to be particularly effective
at inducing rules with labels as conditions in the bodies of the rules.
The resulting models turned out to be indeed very
useful for discovering interesting aspects of the data, which a conventional single-label rule learner is
unable to uncover. The visualizations shown above also confirm that numerous
explicit local and global dependencies can be found in these database.
However, we also found that the \ds{genbase} dataset exhibits only very weak label
dependencies, which can hardly be exploited in order to improve the predictive
performance, despite the fact that this dataset is frequently used for evaluating
multi-label algorithms.

\subsection{Case Study 2: Multi-Label Heads}
\label{sec:casestudy2}

The second case study compares \learner{BR}, \learner{Single} and \learner{Multi}
for candidate rule selection \citep{MLC:Rules-AntiMonotonicity}.\footnote{The source code of the employed algorithms and more extensive evaluations are available at \url{https://github.com/keelm/SeCo-MLC}}
Its main purpose was
to demonstrate the applicability of a covering approach for inducing multi-label head rules despite the exponentially large search space.

\begin{figure}[!t]
	\begin{minipage}{\textwidth}
		\footnotesize
		\fullrule
		\labelfont{red}, \labelfont{green}, \labelfont{blue}, \labelfont{yellow}, \labelfont{white} $\gets$ colors$>$5, stripes$\leq$3 \hfill (65,0) \\
		\labelfontinv{red}, \labelfont{green}, \labelfontinv{blue}, \labelfont{yellow}, \labelfont{white}, \labelfontinv{black}, \labelfontinv{orange}  $\gets$ animate, stripes$\leq$0, crosses$\leq$0 \hfill  (11,0) \\
		{\noindent\rule{1.0\textwidth}{0.3pt}}\\
		\begin{tabular}{p{0.25\textwidth} p{0.1\textwidth} | p{0.25\textwidth} p{0.1\textwidth} | c}
			\labelfont{yellow} $\gets$ colors$>$4      & (21,0) &
			    \labelfont{green} $\gets$ text                  & (11,0) &
			   \multirow{4}{*}{\ \ \resizebox{0.21\textwidth}{!}{\includegraphics{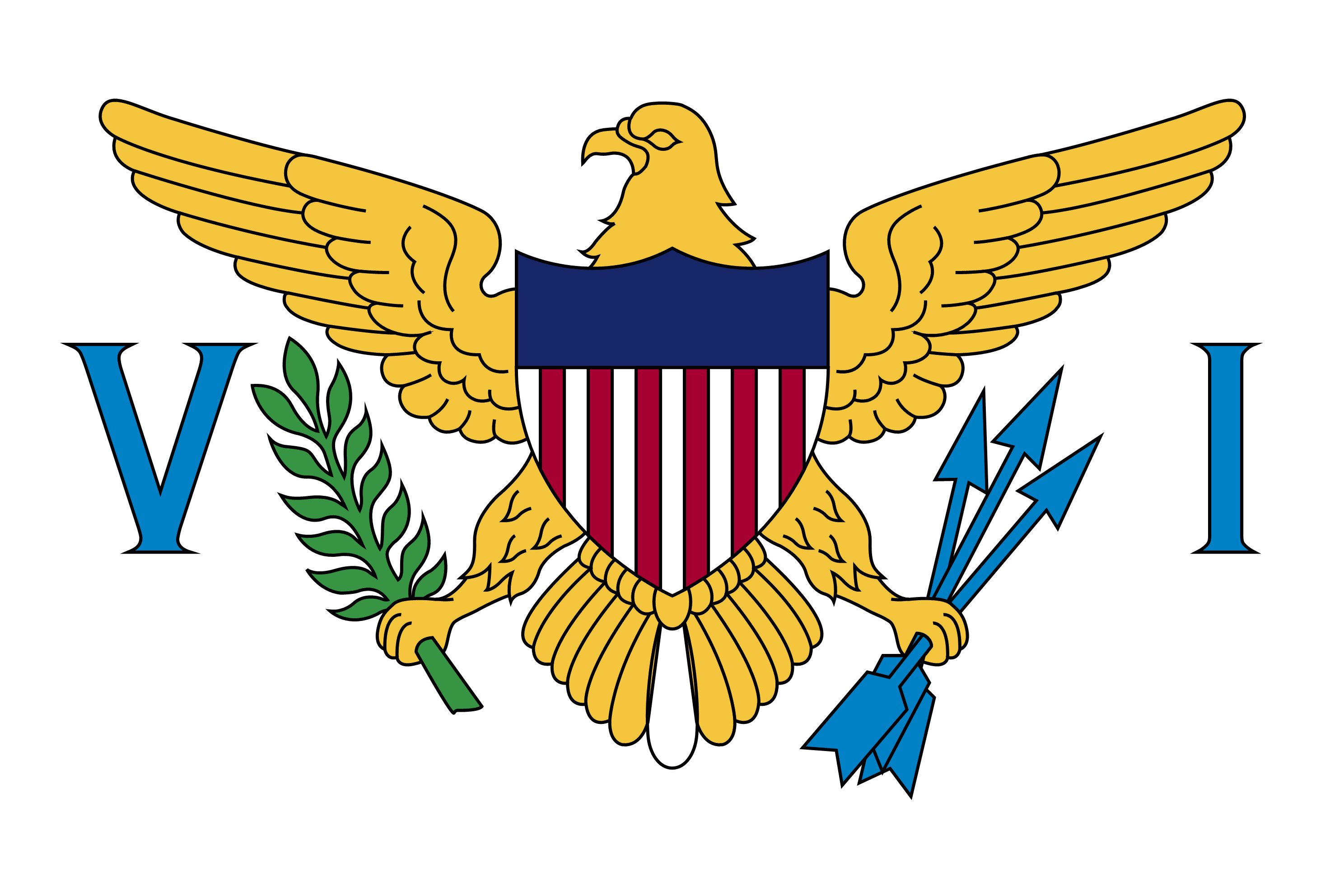}}}
			   			    \\
			\labelfont{red} $\gets$ \labelfont{yellow} & (21,0) &                \labelfontinv{orange} $\gets$ saltires$<$1 & \ (1,0) & \\
			\labelfont{blue} $\gets$ colors$>$5 & (14,0) &
			   \labelfontinv{black} $\gets$ area$<$11 &  (12,0) & \\
			\labelfont{white} $\gets$ \labelfont{blue} & (14,0)& & & \\
		\end{tabular}
		\fullrule
	\end{minipage}
	\caption{Example of learned multi- and single-label head rule lists learned in the \ds{flags} dataset. In parentheses, we show $(TP,FP)$, the number of positive and negative examples covered by each rule. Shown are all rules that cover the flag of the US Virgin Islands, which is shown in the lower right corner.}
	\label{fig:flags}
\end{figure}

\subsubsection{Exemplary Rule Models}

The extended expressiveness of multi-label head rules can be illustrated by
the rules shown in Figure~\ref{fig:flags} that have been learned on the data set \ds{flags}, which maps characteristics of a flag and its corresponding country to the colors appearing on the flag.
Shown are all rules that concern the flag of the US Virgin Islands, which is also shown in the table. Whereas in this case the single-label heads allow an easier visualization of the pairwise dependencies between characteristics/labels and labels, the multi-label head rules allow to represent more complex relationships and provide a more direct explanation of why the respective colors are predicted for the flag.
%
Note that the rules form decision lists, which are applied in order until all labels are set, and later rules cannot overwrite earlier rules. Thus the first rule sets the colors \labelfont{red}, \labelfont{green}, \labelfont{blue}, \labelfont{yellow}, and  \labelfont{white}, whereas the second rule determines that \labelfont{black} and \labelfont{orange} do not occur. The other labels are already set by the previous rule and are not overwritten. No further rules would be considered for the prediction because all labels are already assigned.

This example also illustrates that while decision lists are conceptually easy to extend to the multi-label case by removing covered labels, the interpretability of the resulting rules may suffer. Learning rule sets that collectively determine the predicted label set from multiple possibly overlapping or contradicting partial predictions is an open question for future work.

Whether more labels in the head are more desirable or not highly depends on the data set at hand, the particular scenario and the preferences of the user, as generally do comprehensibility and interpretability of rules.
These issues cannot be solved by the presented methods. 
However,
the flexibility of being able to efficiently find loss-minimizing multi-label heads for a variety of loss functions
can lay the foundation to further improvements, gaining better control over the characteristics of the induced model and hence better adaption to the requirements of a particular use case.

When analyzing the general characteristics of the models which have been learned
by the proposed algorithm, it becomes apparent that multi-label head rules are particularly learned when using the precision metric, rather than one of the other metrics.
The reason is that precision only considers the covered examples whereas for the other metrics the performance also depends on uncovered examples.
Hence, it is very likely that the performance of a rule slightly decreases when adding an additional label to its head, which in turn
causes single-label heads to be preferred.\footnote{The inclusion of a factor which takes the head's size in account could resolve this bias and lead to heads with more labels, but this is subject to future work.}

\subsubsection{Predictive Performance}

Because of this bias towards single-label rules for most of the metrics, large differences in
predictive performance of single-label and multi-label head decision lists cannot be expected.
%
%
%
We therefore
only summarize the main finding, which compared the algorithms' performance using a Friedman test \citep{stats:FriedmanTest} and a Nemenyi post-hoc test \citep{stats:Nemenyi-Test} following the methodology described by \citet{Multiple-Comparisons}.
The null hypothesis of the Friedman test ($\alpha=0.05$, $N=8$, $k=10$) that all 10 algorithms have the same predictive quality on the eight datasets shown in Figure~\ref{tab:datastats} (excluding \ds{enron})
could not be rejected for many of the evaluation measures, such as subset accuracy and micro- and macro-averaged F1. In the other cases, the Nemenyi post-hoc test was not able to assess a statistical difference between different algorithms that used the same objective for optimizing the rules.

\begin{figure}[t!]
  \centering
  \includegraphics[width=0.45\textwidth]{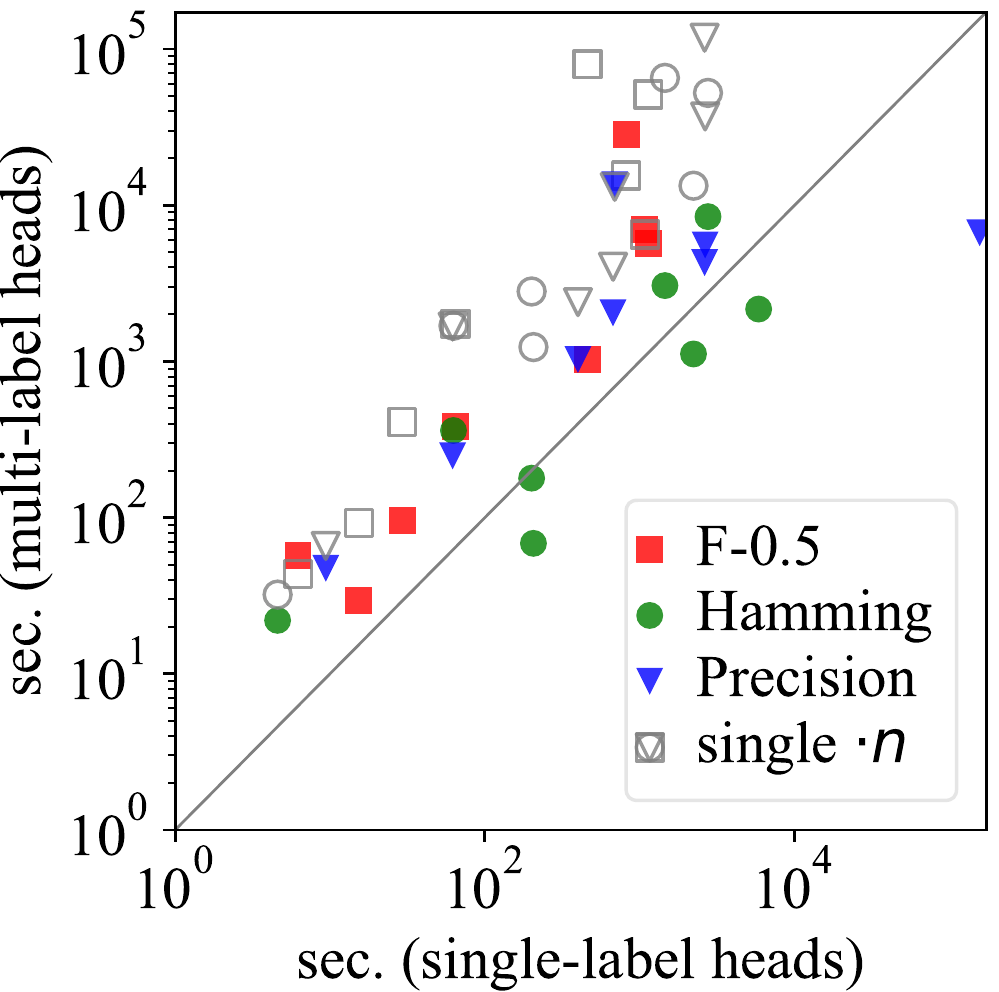}
  \caption{Training times for the separate-and-conquer algorithm. Direct comparison between learning single-label and multi-label heads.}
  \label{fig:time_comp}
\end{figure}

\subsubsection{Computational Cost}
Figure~\ref{fig:time_comp} shows the relation between the time spent for finding single- vs. multi-label head rules using the same objective and data set.
The empty forms denote the single-label times multiplied by the number of labels in the data set and represent an approach with a computational complexity increased by one polynomial order w.r.t. number of labels.
Note that full exploration of the labels space was already intractable for the smaller data sets on our system, and became only feasible through the use of anti-monotonicity and decomposability pruning, as described in Section~\ref{sec:loss-minimization}. We can observe that the costs for learning multi-label head rules are in the same order of magnitude as the costs for learning single-label head rules, despite the need for exploring the full label space for each candidate body.


\section{Conclusion}
\label{sec:conclusions}

In this work, we recapitulated recent work on inductive rule learning for multi-label classification problems. The main advantage of such an approach is that mixed dependencies between input variables and labels can be seamlessly integrated into a single coherent representation, which facilitates the interpretability of the learned multi-label models.
However, we have also seen that combining multi-label rules into interpretable predictive theories faces several problems, which are not yet sufficiently well addressed by current solutions.
One problem is that mixed-dependency rules needs to be structured in a way that allows each label that occurs in the body of a rule to be predicted by some other rule in a way that avoids cyclic reasoning. We have seen two principal approaches to solve this problem, a layered technique that relies on
a pre-defined structure of the prediction and rule induction process, 
 and a second approach that relies on adapting the separate-and-conquer or covering strategy from single-label rule learning to the multi-label case.
The results we have shown in several domains are encouraging, but it is also clear that they are still somewhat limited. For example, the multi-label decision lists that result from the latter approach are hard to interpret because of the implicit dependencies that are captured in the sequential interpretation of the rules.
Thus, multi-label rule learning remains an interesting  research goal, which combines challenging algorithmic problems with a strong application potential.


\begin{acknowledgement}
We would like to thank Frederik Janssen for his contributions to this work. Computations for this research were conducted on the Lichtenberg high performance computer of the TU Darmstadt.
\end{acknowledgement}

\def\bibfont{\footnotesize} 
\bibliographystyle{splncsnat}
\bibliography{literatur-all,references,rules,jf,ilp}

\end{document}